\documentclass[10pt,twocolumn,letterpaper]{article}

\usepackage{cvpr}              

\usepackage[utf8]{inputenc} 
\usepackage[T1]{fontenc}    
\usepackage{url}            
\usepackage{booktabs}       
\usepackage{amsfonts}       
\usepackage{nicefrac}       
\usepackage{microtype}      
\usepackage{xcolor}         
\usepackage{graphicx}
\usepackage{amsmath}

\usepackage{multirow,tabularx}
\usepackage[table]{xcolor}
\usepackage{enumitem}
\usepackage{xspace}
\usepackage{pifont}
\definecolor{cayenne}{RGB}{220, 53, 69}         
\definecolor{espresso}{RGB}{147, 68, 78}    
\newcommand{\xmark}{\textcolor{red}{\ding{55}}}
\newcommand{\modelname}{\textbf{\textcolor{espresso}{P}\textcolor{cayenne}{art}$^{\textcolor{cayenne}{2}}$\textcolor{espresso}{GS}\xspace}}

\newcommand{\modelnamenc}{{Part$^{2}$GS}\xspace}

\definecolor{cvprblue}{rgb}{0.21,0.49,0.74}
\usepackage[pagebackref,breaklinks,colorlinks,allcolors=cvprblue]{hyperref}

\def\paperID{} 
\def\confName{CVPR}
\def\confYear{2026}

\definecolor{IllinoisOrange}{HTML}{FF5F05}
\definecolor{IllinoisBlue}{HTML}{13294B}
\newcommand{\logoicon}{%
  \raisebox{-0.20\height}{\includegraphics[height=1.2em]{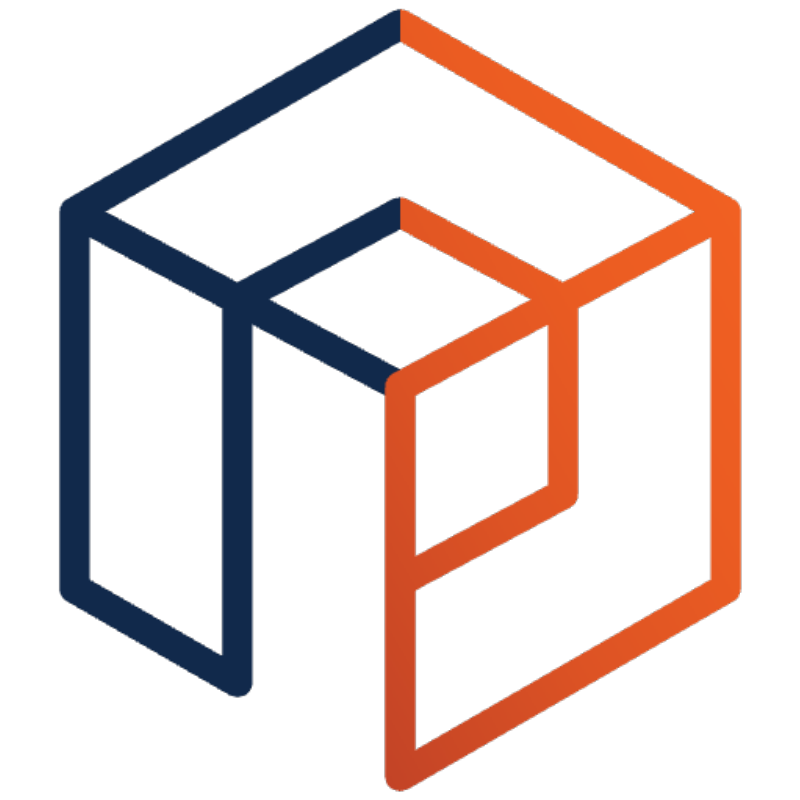}}%
}

\title{\textcolor{espresso}{P}\textcolor{cayenne}{art}-aware Modeling of \textcolor{cayenne}{Art}iculated Objects using 3D \textcolor{espresso}{G}aussian \textcolor{espresso}{S}platting}

\author{
Tianjiao Yu \quad
Vedant Shah \quad
Muntasir Wahed \quad
Ying Shen \quad
Kiet A. Nguyen \quad
Ismini Lourentzou\\
University of Illinois Urbana-Champaign\\
{\tt\small \{ty41, vrshah4, mwahed2, ying22, kietan2, lourent2\}@illinois.edu}
}

\begin{document}
\maketitle

\begin{abstract}
Articulated objects are common in the real world, yet modeling their structure and motion remains a challenging task for 3D reconstruction methods. 
In this work, we introduce \modelname{}, a novel 3D Gaussian splatting framework for modeling articulated digital twins of multi-part objects with high-fidelity geometry and physically consistent articulation. 
\modelnamenc{} augments each Gaussian with a learnable part-identity embedding and learns a motion-aware canonical representation that encodes physical constraints such as contact, velocity consistency, and vector-field alignment. 
To ensure collision-free motion, we introduce a repel-point field that stabilizes joint trajectories and enforces realistic part separation. 
Experiments across several benchmarks, covering a wide range of articulation types, show that \modelnamenc{} consistently outperforms state-of-the-art methods by up to 10$\times$ in Chamfer Distance for movable parts.\looseness-1 

\noindent \logoicon~\href{https://plan-lab.github.io/part2gs}{\textcolor{IllinoisBlue}{PLAN Lab}~\textcolor{IllinoisOrange}{https://plan-lab.github.io/part2gs}}
\end{abstract}

\section{Introduction}
Articulated objects are ubiquitous in our physical world and central to interaction and manipulation tasks.
Creating faithful 3D assets of such objects is valuable for a variety of applications in 3D perception ~\cite{deitke2023objaverse, deng2023banana, geng2023gapartnet, liu2023semi, liu2023self, liu2025medsam3, yu2026dreampartgen, yu2025core3d}, embodied AI ~\cite{deitke2022, kolve2017ai2, puig2023habitat, yu2025uncertainty,shen2025elba}, and robotics ~\cite{gadre2021act,mo2021where2act,qian2023understanding, liu2026palm}. 
Despite their utility, most available articulated 3D assets are created manually, and existing datasets are often limited in both scale and diversity ~\cite{jain2021screwnet, liu2023paris, liu2022akb}, restricting advancements in intelligent systems that can effectively understand and manipulate articulated objects in diverse, real-world environments. To address this challenge, recent efforts have focused on reconstructing articulated objects from real-world observations~\cite{heppert2023carto, song2024reacto, shen2025gaussianart} or predicting articulation patterns for existing 3D models~\cite{lei2023nap,liu2024cage, wu2025reartgs}. However, these methods often rely on labor-intensive data collection processes or large, predefined datasets of 3D objects with detailed geometry.

Recent advances in articulated 3D object reconstruction have leveraged 3D Gaussian Splatting (3DGS) and Neural Radiance Fields (NeRFs) to model object geometry and motion from visual observations~\cite{guo2025articulatedgs,liu2025building,song2024reacto,swaminathan2024leia}. Despite their effectiveness, these approaches largely treat articulation as a geometric interpolation problem, without incorporating physical feasibility or semantic part understanding. As a result, they often produce reconstructions that are not well grounded in object mechanics, exhibiting artifacts such as floating components or physically implausible joint behavior, particularly for complex multi-part objects. Moreover, existing methods rely heavily on direct state-to-state interpolation and clustering, which do not enforce rigid-body consistency or articulation constraints in unconstrained settings~\cite{le2024articulate,liu2025building}.

\begin{figure}[t]
  \centering
  \includegraphics[width=0.95\linewidth]{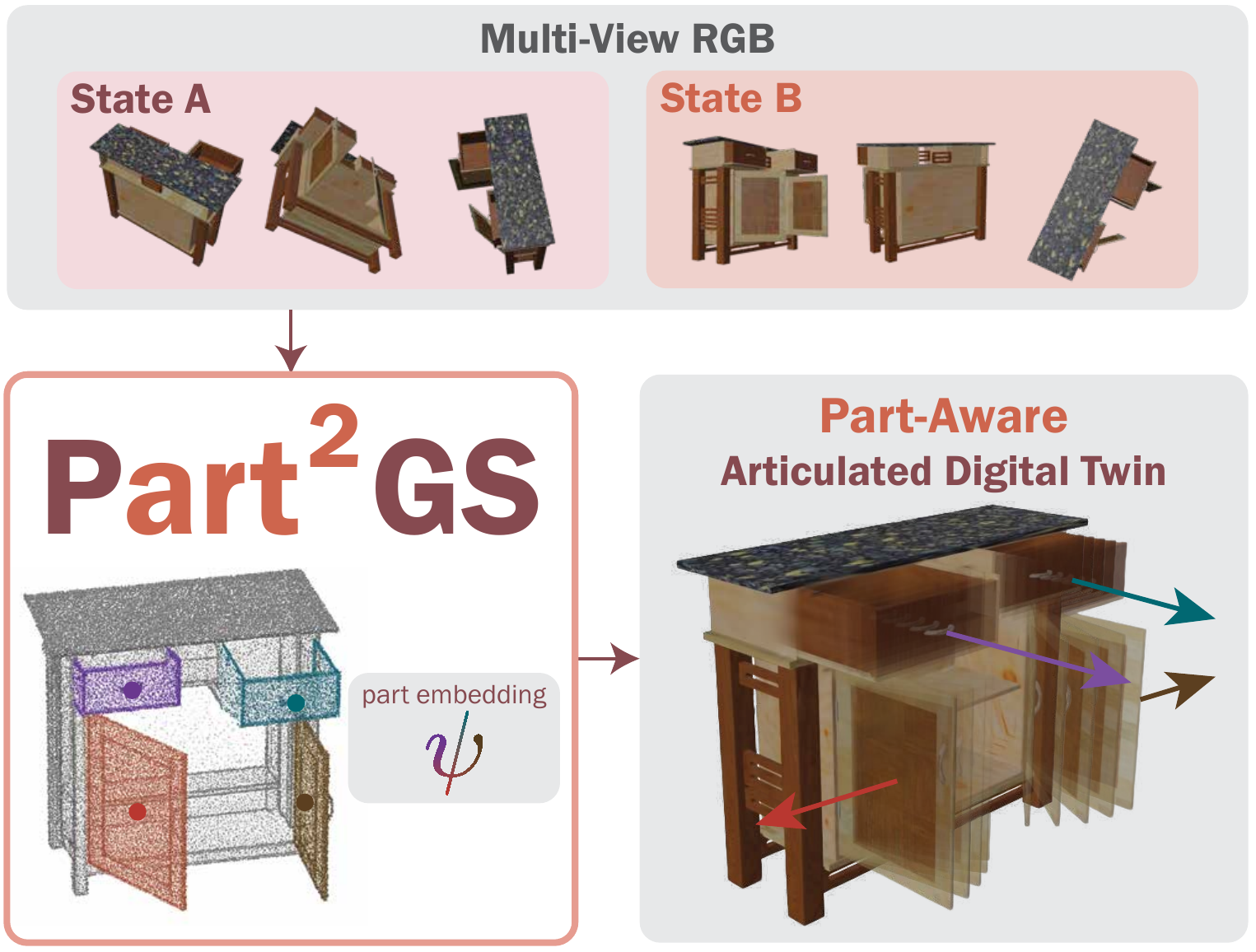}
  \vspace{-0.3cm}
  \caption{\textbf{\modelname{} reconstructs articulated 3D objects from multi-view observations.} Our method augments each Gaussian with a learnable part-identity embedding
  that allows part structure to emerge directly from geometry, motion, and physical constraints.}  
  \label{fig:teaser}
\end{figure}
To overcome these limitations, we introduce \textbf{\textcolor{espresso}{P}\textcolor{cayenne}{art}-aware Object \textcolor{cayenne}{Art}iculation with 3D \textcolor{espresso}{G}aussian \textcolor{espresso}{S}platting} (\textbf{\modelname}), a novel part-disentangled, physics-grounded framework for reconstructing articulated 3D digital twins from raw multi-view observations. \modelnamenc models object parts as learnable Gaussian attributes, which are coupled with motion-aware canonicalization and physics-informed articulation learning, enabling recovery of both high-fidelity geometry and physically coherent motion.

\modelnamenc addresses three core challenges:
\textbf{\ding{182} Unstructured Part Articulation:} Rather than relying solely on unsupervised clustering, dual-quaternion blending, or using predefined part ground truth, \modelnamenc{} introduces a part parameter into the standard Gaussian parameters, and guides part transformation with physics-aware forces and learned part embeddings. This allows emergent, differentiable part discovery that aligns geometric and kinematic structure. To further ensure inter-part separation, we introduce a field of repel points that apply localized repulsive forces at contact regions, guiding parts toward smooth and physically valid motion trajectories.
\textbf{\ding{183} Lack of Physical Constraints:} Existing methods lack grounding, collision avoidance, and coherent rigid-body motion, resulting in implausible part behavior ~\cite{liu2024cage,liu2024singapo}. 
\modelnamenc{} integrates physically motivated losses such as contact constraints, velocity consistency, and vector-field alignment to enforce grounded, collision-free, realistic articulation. \textbf{\ding{184} Rigid State-Pair Modeling:} Prior methods rely heavily on fixed, geometric interpolation between two states ~\cite{liu2023paris,liu2025building,weng2024neural}. In contrast, \modelnamenc{} builds a motion-aware canonical representation that adaptively biases interpolation toward the more informative, motion-rich state via a learnable coefficient, leading to better part disentanglement.\looseness-1
 
Through extensive experiments, we demonstrate that \modelnamenc{} achieves state-of-the-art performance in reconstructing articulated 3D objects, delivering high-fidelity geometry and physically consistent motion, even in challenging multi-part scenarios. Our contributions are summarized as follows:
\begin{itemize}
    \item We introduce \textbf{\modelname}, a part-aware 3D Gaussian framework for articulated object reconstruction that jointly optimizes geometry, part discovery, and physically consistent articulation from raw multi-view observations.
    \item We propose a \emph{motion-aware canonical representation} with physics-informed articulation and a novel \emph{repel-point mechanism} that applies localized repulsive forces at part boundaries, to produce part-disentangled geometry with smooth, collision-free, physically consistent articulation.     
    \item We extensively evaluate \modelnamenc across diverse articulated objects and benchmarks, showing consistent state-of-the-art performance over strong baselines, with substantial gains in articulation accuracy and reconstruction quality.
\end{itemize}

\section{Related Work}
\subsection{Articulated Object Modeling} 
Early work on articulated object modeling relied primarily on geometric reasoning and hand-crafted heuristics. Given a mesh, slippage analysis and probing techniques were used to detect rotational and translational axes by observing when two parts penetrate or slip past each other~\citep{xu2009joint}, and joint types and limits were set by trial‐and‐error bisection~\citep{li2016mobility,mitra2010illustrating,sharf2014mobility}. 
More recent supervised approaches learn canonical object- and part-level coordinate spaces, to map arbitrary poses to a template frame, then recover joints by fitting rigid transforms~\citep{geng2023gapartnet,hu2017learning,li2020category}.  To reduce reliance on labeled data, self-supervised methods replace labels with correspondence- or reconstruction-based objectives. Some infer articulation by tracking points across frames and fitting motion trajectories~\citep{shi2021self}, while single-image methods recover joint transformations by warping parts to and from learned canonical spaces~\citep{liu2023paris,liu2023self}. 

Despite these advances, such methods rely on external structural priors, such as predefined part libraries, kinematic graphs, or category-specific templates~\citep{jiang2022ditto,lei2023nap,liu2024cage,liu2024singapo}. In contrast, \modelnamenc{} recovers part decompositions and articulation parameters directly from raw multi-view observations.

\subsection{Dynamic Gaussian Modeling} 
Building on the seminal 3D Gaussian Splatting framework~\cite{kerbl20233dgaussiansplattingrealtime}, a broad body of follow-up work has extended Gaussian representations to dynamic and 4D settings. Prior methods model temporal variation through per-Gaussian deformation fields for animatable human avatars~\cite{jung2023deformable} or by smoothly evolving Gaussian attributes over time to replay dynamic scenes~\cite{wu20244d}.
Other approaches improve temporal coherence and geometric fidelity by preserving Gaussian identities across frames, introducing temporal features for live novel-view rendering, or constraining deformations to respect local surface geometry~\citep{li2024spacetime,lu20243d,luiten2024dynamic,vilesov2023cg3d}. 

A related line of work targets animatable avatars and scenes, learning per-splat pose controls, disentangling motion modes, or removing the need for predefined templates~\citep{bae2024per,qian20243dgs,wan2024template}. In parallel, sparse superpoint-based formulations enable direct and interactive editing of Gaussian groups in real time, prioritizing user-controllable deformability over recovery of physical or kinematic structure~\citep{huang2024sc,wan2024superpoint}. 

Despite these advances, existing methods are primarily designed for continuous non-rigid deformation, such as soft-body dynamics or general scene flow, rather than part-based articulated motion~\cite{ye2024gaussian,zhang2024magicpose4d,wu20244d,li2024st,gao2024gaussianflow}. We introduce a part-aware dynamic Gaussian modeling framework that explicitly links motion to automatically discovered part structure, enabling fine-grained and physically grounded articulation.
\begin{figure*}[t!]
    \centering
    \includegraphics[width=0.99\linewidth]{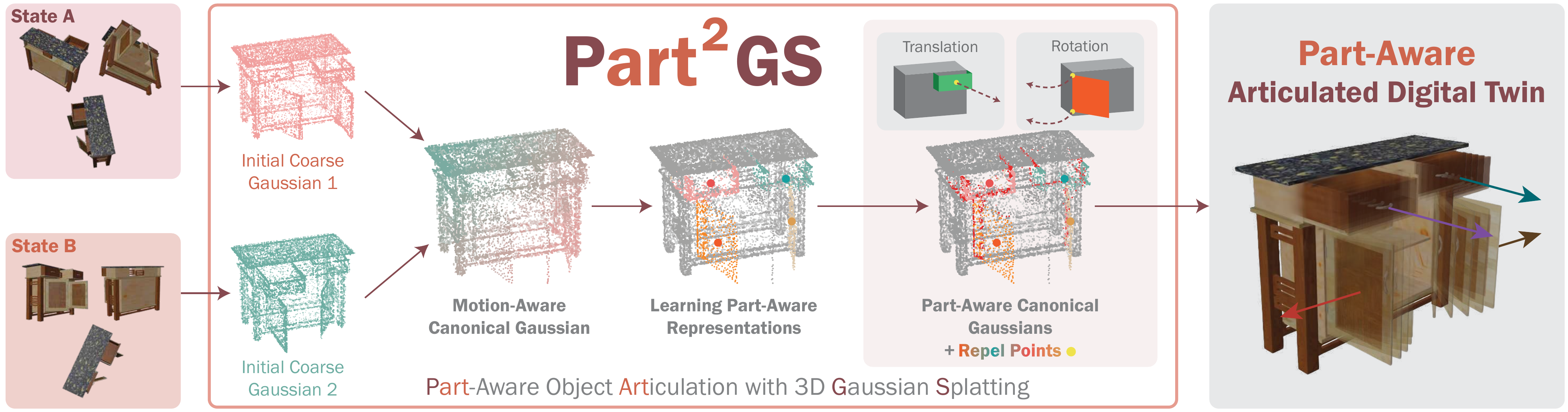}
    \vspace{-0.3cm}
    \caption{\textbf{\modelname{} Overview.} \modelnamenc{} reconstructs articulated 3D objects as part-aware digital twins from multi-view observations across different states. \modelnamenc{} first initializes coarse 3D Gaussian fields and aligns them into a shared motion-aware canonical space. 
    Part-aware representations are subsequently learned through per-Gaussian part embeddings and physics-guided regularization, enabling each part’s translation and rotation to be disentangled from overall deformation.
    Finally,  \modelnamenc{} optimizes part-level SE(3) motions with repel-point fields and physical constraints, producing accurate part boundaries and collision-free articulation.}
    \label{fig:main_method}
\end{figure*}
\section{Preliminaries}
\label{sec:prelim}
\textbf{3D Gaussian Splatting.} 3D Gaussian Splatting~\cite{kerbl20233dgaussiansplattingrealtime} (3DGS) is a  state-of-the-art approach for representing 3D scenes by parameterizing them as collections of anisotropic Gaussians. Unlike implicit representation methods such as
NeRF \cite{mildenhall2021nerf}, which relies on volume rendering, 3DGS achieves real-time rendering by splatting these Gaussians onto a 2D plane and compositing their effects through differentiable alpha blending \cite{yifan2019differentiable}. Formally, a scene is modeled as a set of $N$ anisotropic Gaussians, denoted as
\begin{equation}
\setlength{\abovedisplayskip}{7pt}
\setlength{\belowdisplayskip}{7pt}
    \mathcal{G} = \{G_i : \boldsymbol{\mu}_i, \boldsymbol{r}_i, \boldsymbol{s}_i, \sigma_i, \boldsymbol{h}_i\}_{i=1}^{N},
    \label{eq:g_params}
\end{equation}
where each Gaussian $G_i$ is parameterized by its centroid position $\boldsymbol{\mu}_i \in \mathbb{R}^3$, rotation quaternion $\boldsymbol{r}_i \in \mathbb{R}^4$, anisotropic scale vector $\boldsymbol{s}_i \in \mathbb{R}^3$, scalar opacity $\sigma_i \in [0,1]$, and spherical harmonics coefficients $\boldsymbol{h}_i$ that encode view-dependent appearance. The opacity value of a Gaussian $G_i$ at any spatial point $\boldsymbol{x}\in\mathbb{R}^3$ is computed as
\begin{equation}
\setlength{\abovedisplayskip}{7pt}
\setlength{\belowdisplayskip}{7pt}
    \alpha_i(\boldsymbol{x}) = \sigma_i \exp\left(-\frac{1}{2}(\boldsymbol{x}-\boldsymbol{\mu}_i)^\top \boldsymbol{\Sigma}_i^{-1}(\boldsymbol{x}-\boldsymbol{\mu}_i)\right).
\end{equation}
The covariance matrix $\boldsymbol{\Sigma}_i$ characterizing the anisotropic spread of the Gaussian is defined as $\boldsymbol{\Sigma}_i = \boldsymbol{R}_i \boldsymbol{S}_i \boldsymbol{S}_i^\top \boldsymbol{R}_i^\top.$
Here, $\boldsymbol{S}_i$ is a diagonal matrix of scaling factors, and $\boldsymbol{R}_i$ is a rotation matrix corresponding to quaternion $\boldsymbol{r}_i$. This decomposition ensures that the covariance matrix remains positive semi-definite, maintaining a valid geometric interpretation of Gaussian spread and orientation.
To render a scene, each Gaussian is projected onto the image plane and composited through differentiable $\alpha$-blending, which accumulates their opacity and spherical harmonic–based color contributions. Formally, the rendered image $\boldsymbol{I}$ is expressed as\looseness-1
\begin{equation}
\setlength{\abovedisplayskip}{7pt}
\setlength{\belowdisplayskip}{7pt}
\boldsymbol{I}\!=\!\sum_{i=1}^{N} T_i\, \alpha_i^{\mathbb{R}^2}\, \mathcal{H}(\boldsymbol{h}_i, \boldsymbol{v}_i), 
\text{ where } T_i\!=\!\prod_{j=1}^{i-1}(1-\alpha_j^{\mathbb{R}^2}).
\end{equation}
Here, $\alpha_i^{\mathbb{R}^2}$ is the projected 2D Gaussian opacity evaluated at each pixel coordinate, analogous to its 3D counterpart. The term $\mathcal{H}(\boldsymbol{h}_i,\boldsymbol{v}_i)$ represents the spherical harmonics-based color function evaluated along viewing direction $\boldsymbol{v}_i$, while the blending weights $T_i$ encode front-to-back occlusion and transparency effects. Given $N$ multi-view images $\mathcal{I}\!=\!\{\boldsymbol{I}_i\}_{i=1}^{N}$, the Gaussian parameters $\mathcal{G}$ are optimized by minimizing a differentiable rendering loss
\begin{equation}
\setlength{\abovedisplayskip}{7pt}
\setlength{\belowdisplayskip}{7pt}
    \mathcal{L}_{\text{render}} = (1 - \lambda)\mathcal{L}_I + \lambda\mathcal{L}_{\text{D-SSIM}},
    \label{eq:render_loss}
\end{equation}
where $\mathcal{L}_I = ||\boldsymbol{I} - \boldsymbol{{I}^{*}}||_1$ is the pixel-wise $\ell_1$ reconstruction loss, $\mathcal{L}_{\text{D-SSIM}}$ measures perceptual structural similarity between rendered and target images~\citep{kerbl20233dgaussiansplattingrealtime}, and $\lambda$ is the loss coefficient.  This explicit Gaussian-based scene representation, combined with a differentiable rendering process, enables efficient inference of the 3D structure directly from view-based supervision. 
\section{\modelname: \textcolor{espresso}{P}\textcolor{cayenne}{art}-aware Object \textcolor{cayenne}{Art}iculation} 
\label{sec:method_main}
In this work, we introduce {\modelnamenc}, a method that constructs articulated 3D object representations by leveraging 3D Gaussian Splatting for part-aware geometry and articulation learning. Given a set of 2D multi-view images $\mathcal{I}_t\!=\!\{\boldsymbol{I}_i^t\}_{i=1}^{N}$ captured at two distinct joint states $t\in \{0,1\}$, our objective is to generate an articulated 3D object representation $\mathcal{O}$ with part-level disentanglement and physically grounded motion. $\mathcal{O}$ is modeled as a composition of a static base $\mathcal{G}_{\text{static}}$ and $K$ movable parts, represented as $\mathcal{G}\!=\!\{\mathcal{G}_{\text{static}}, \mathcal{G}_{k} \mid k \in [1, \dots, K]\}$. Each part $\mathcal{G}_k$ is modeled as a collection of $M_k$ 3D Gaussians $\mathcal{G}_k\!=\!\{G^k_{i} \mid i \in [1,\dots,M_k]\}$, enabling flexible manipulation and clear part delineation.

As illustrated in \Cref{fig:main_method}, \modelnamenc{} constructs a \textit{motion-aware canonical Gaussian field} by aligning and merging single-state reconstructions from two joint configurations, $\mathcal{I}_0$ and $\mathcal{I}_1$ (\S\Cref{sec:part}). Each Gaussian $\mathcal{G}_i$ is augmented with a compact, learnable \textit{part-identity embedding} $\boldsymbol{\psi}_i$ that enables unsupervised grouping into physically coherent parts (\S\ref{sec:partdisc}). The motion of each discovered part is modeled as an $\mathrm{SE}(3)$ rigid transformation. To ensure collision-free articulation, \modelnamenc{} introduces \textit{repel points} along part interfaces that generate localized repulsive potentials, stabilizing joint trajectories and preventing interpenetration (\S\ref{sec:articulate}). Finally, \textit{physics-informed regularization} constrains each part to follow consistent, rigid-body dynamics, yielding stable and physically plausible articulation (\S\ref{sec:phys}).\looseness-1

\subsection{Motion-Aware Canonical Gaussian}
\label{sec:part}
Prior approaches that rely on directly modeling correspondences between two distinct states often suffer from severe occlusion, viewpoint inconsistencies, and difficulties arising from learning articulation deformation while maintaining rigid geometry~\cite{jiang2022ditto, weng2024neural}.
To overcome these limitations, we construct a motion-aware canonical Gaussian field that adaptively fuses the two single-state reconstructions.
We first establish correspondences between $\mathcal{G}^0_{\text{single}}$ and $\mathcal{G}^1_{\text{single}}$ via Hungarian matching based on pairwise distances between Gaussian centers. For each matched pair, rather than simply averaging \cite{liu2025building}, we create a canonical Gaussian by interpolating between the two corresponding Gaussians. 

Specifically, we introduce a \textit{motion-informed prior} to guide the interpolation. We estimate the motion richness of each state by computing the mean minimum distance from each Gaussian in one state to its nearest neighbor in the other state. Formally, for each state $t \in \{0,1\}$, we compute
\begin{equation}\label{eq:D}
\setlength{\abovedisplayskip}{6pt}
\setlength{\belowdisplayskip}{6pt}
\text{D}^{t \to \bar{t}}
= \mathbb{E}_i \!\left[
\min_{j}
\left\|
\boldsymbol{\mu}_i^{(t)}
- \boldsymbol{\mu}_j^{(1 - t)}
\right\|_2
\right], 
\end{equation}
where $\bar{t} = 1 - t$ denotes the opposite state.
The state with the higher $\text{D}^{t \to \bar{t}}$ value is identified as the \textit{motion-informative state}, reflecting greater articulation or part displacement. 
For a matched Gaussian pair $(G_i^0, G_i^1)$, the canonical Gaussian $G_i^c$ is computed as $\boldsymbol{\mu}_i^c = \beta \boldsymbol{\mu}_i^0 + (1 - \beta) \boldsymbol{\mu}_i^1$,
where $\beta\!=\!\frac{D_{0\rightarrow1}}{D_{0\rightarrow1}+D_{1\rightarrow0}} \in [0,1]$ is adaptive weighting coefficient determined by the relative motion richness scores $D_{0\rightarrow1}$ and $D_{1\rightarrow0}$ as defined in \Cref{eq:D}.

\subsection{Learning Part-Aware Representations}
\label{sec:partdisc}
To achieve a detailed and controllable representation of articulated objects, it is crucial to explicitly model the object’s semantic decomposition into parts.  While the standard 3D Gaussian Splatting approach provides efficient geometric reconstruction, it lacks explicit part-level semantics necessary for articulated object modeling. Motivated by this, we augment each Gaussian representation, introduced in ~\cref{eq:g_params},
with a compact, learnable \textit{part-identity embedding} $\boldsymbol{\psi}_{i}$ that encodes latent part membership and geometric affinity.

To ensure that neighboring Gaussians on the same surface receive consistent part assignments, we impose a neighborhood-consistency regularization loss that enforces 3D spatial consistency by encouraging similar encodings among neighboring Gaussians:
\begin{equation}
\setlength{\abovedisplayskip}{8pt}
\setlength{\belowdisplayskip}{8pt}
\scalebox{0.95}{$
\mathcal{L}_{\text{part}}\!=\!\frac{1}{M} \sum\limits_{i=1}^{M} D_{\text{KL}}\left(F(G_i)\,\Big|\Big|\, \frac{1}{|\mathcal{N}(G_i)|} \sum\limits_{j \in \mathcal{N}(G_i)} F(G_j) \right),
$}
\end{equation}
where $M$ is the number of Gaussians in the current batch, $F(G_i)\!=\!\text{softmax}(f(\boldsymbol{\psi}_i))$ is the part identity probability distribution for each Gaussian $G_i$, computed by projecting part-identity encodings into $K$ part categories through a shared linear layer $f$ followed by a softmax operation, and $\mathcal{N}(G_i)$ denotes the k-nearest neighbors in 3D space computed based on the L2 distance between Gaussian centers.

\begin{figure*}[t!]
  \centering
  \includegraphics[width=0.99\textwidth]{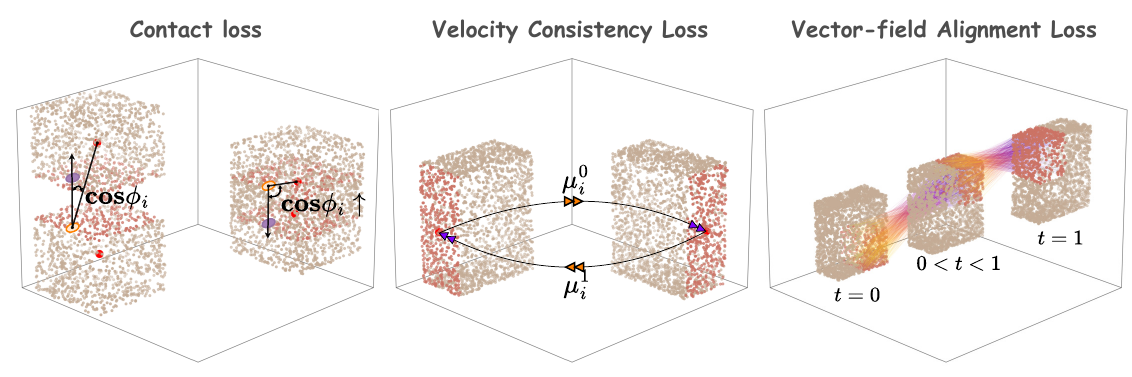}
  \vspace{-0.5cm}
  \caption{\textbf{Physics-Informed regularization constraints}. (1) \textit{Contact Loss} penalizes interpenetration by minimizing the angle between two vectors for each Gaussian: a) the vector pointing to the center of the opposing part, and b) the vector pointing to its nearest Gaussian in that part. The red dots (\textcolor{red}{$\boldsymbol{\bullet}$}) denote the object centers. (2) \textit{Velocity Consistency} encourages similar displacement vectors within each rigid part (\eg $\mu_i^0 == \mu_i^1$). Red dots (\textcolor{red}{$\boldsymbol{\bullet}$}) represent the same Gaussian at different states. (3) \textit{Vector-field Alignment } enforces consistency between predicted part transformations and observed motions (\S\ref{sec:phys}).\looseness-1}
  \label{fig:physical_constraints}
\end{figure*}

\label{sec:physical_constraints}
    
\subsection{Repulsion-Guided Articulation Optimization}
\label{sec:articulate}
To enable realistic articulation of the object’s movable parts relative to its static base, we introduce \textit{repel points}, $\mathcal{R}=\{\mathbf{r}_j \in \mathbb{R}^3 \mid j=1,2,\ldots,N_R\}$, where $N_R$ is the total number of repel points, and each $\mathbf{r}_j$ 
is associated with a repulsion field that encourages each movable part to find a stable configuration while avoiding excessive overlap with the static base. 
These repel points, placed in regions of articulated parts where the static and movable parts are initially close, apply localized repulsive forces that guide the movable part's movement while maintaining physical separation. 
The repulsion force is defined as 
\begin{equation}\label{eq:repel}
\setlength{\abovedisplayskip}{6pt}
\setlength{\belowdisplayskip}{6pt}
\mathbf{F}^k_{\text{repel}, i} = \sum_{\mathbf{r}_j \in \mathcal{R}} k_r \cdot \frac{(\mathbf{r}_j - \boldsymbol{\mu}^k_i)}{\|\mathbf{r}_j - \boldsymbol{\mu}^k_i\|^3},
\end{equation}
where $k_r$ is a repulsion coefficient, $\boldsymbol{\mu}_i$ is the center of the Gaussian $G_i$, $\mathbf{r}_j$ is the $j$-th repeller point, and $\mathbf{F}^k_{\text{repel}, i}$ is the force vector applied to Gaussian $G^k_i$.

To capture feasible movement trajectories, each movable part undergoes a rigid transformation 
$T_k\!=\!(\mathbf{R}_k, \mathbf{t}_k) \in \mathrm{SE}(3)$, where $R_{k} \in \mathrm{SO}(3)$ is the rotation matrix and $t_{k} \in \mathbb{R}^3$ denotes the translation vector of the $k$-th movable part with respect to the static base. 
To learn the true movement, we initialize with random transformations $T^{(0)}_k\!=\!(\mathbf{R}^{(0)}_k, \mathbf{t}^{(0)}_k)$ and iteratively refine them by aligning the predicted positions of the Gaussian centers with their observed locations during articulation. 
Specifically, at each iteration step $t$, the transformed position of each Gaussian $G_i^k$ under the current transformation is calculated as $\boldsymbol{\mu}_i^{k, (t)}\!=\!\mathbf{R}_k^{(t)} \boldsymbol{\mu}_i^{k, 0} + \mathbf{t}_k^{(t)}$, where $\boldsymbol{\mu}_i^{k, 0}$ is the initial canonical position of the Gaussian.
To enforce collision-free motion, each Gaussian is further adjusted based on the influence of nearby repel points, \ie, $\boldsymbol{\mu}_i^{k, (t)} \leftarrow \boldsymbol{\mu}_i^{k, (t)} + \mathbf{F}_{\text{repel}, i}^k$.

We optimize the part trajectories by minimizing an articulation loss that enforces both positional alignment and rotational consistency at each iteration step $t$, \ie,
\begin{equation}
\setlength{\abovedisplayskip}{6pt}
\setlength{\belowdisplayskip}{6pt}
\begin{aligned}
\mathcal{L}_{\text{art}}^{(t)}\!=\!
\sum_{k=1}^K \sum_{i \in \mathcal{G}_k}
\bigl\| \mathbf{R}_k^{(t)} \boldsymbol{\mu}_i^{k,0} + \mathbf{t}_k^{(t)} + \mathbf{F}_{\text{repel}, i}^k - \hat{\boldsymbol{\mu}}_i^k \bigr\|^2 \\
\hspace{-0.3cm} + \lambda_{\text{rot}}\operatorname{Angle}\!\bigl(\mathbf{R}_{k}^{(t)}, \hat{\mathbf{R}}_{k}
\bigr),
\end{aligned}
\end{equation}
where $\lambda_{\text{rot}}$ is a weighting factor enforcing rotational alignment and $\text{Angle}(\cdot)$ measures the rotational deviation.

Additionally, we leverage the aforementioned contact loss $\mathcal{L}_{\text{contact}}$ and $\mathcal{L}_{\text{part}}$ to prevent the movable part from overlapping with the static base or other parts, ensuring physical plausibility throughout the articulation process. 
Through this iterative process, we converge on a set of transformations $\mathcal{T} = \{T_{k} \mid k \in [1, \dots, K]\}$ that capture realistic movement paths of each movable part with respect to the static base. 

This articulation learning framework, grounded in repel points, transformation refinement, and contact-aware constraints, provides a robust model for representing and manipulating the articulated parts of the object $\mathcal{O}$.

\subsection{Physics-Informed Regularization}\label{sec:phys}
To preserve the physical plausibility of articulated motion, we incorporate three auxiliary losses that constrain part-level deformation: contact loss, vector-field alignment, and velocity consistency (See \Cref{fig:physical_constraints}).

First, the \textit{contact loss} discourages unrealistic interpenetration between movable parts and the static base by introducing a contact-based constraint.
For each Gaussian center $\boldsymbol{\mu}_i \in G_i^k$ belonging to movable part $\mathcal{G}_k$, we locate its nearest corresponding static Gaussian center
$\boldsymbol{\mu}_i^{\star}$. 
Let $\boldsymbol{\bar{\mu}}$ be the centroid of the static base $\mathcal{G}_{\text{static}}$, and define $\mathbf{d}_i\!=\!\boldsymbol{\mu}_i - \boldsymbol{\mu}_i^{\star}, ~~\mathbf{d}_{k}\!=\!\boldsymbol{\mu}_i - \bar{\boldsymbol{\mu}}$, where $\mathbf{d}_i$ represents the offset from the movable part to its nearest static Gaussian, and  $\mathbf{d}_k$ captures the displacement from the movable part to the centroid of the static base.
The cosine of the angle $\varphi_i$ between these two vectors penalizes obtuse contact angles via
\begin{equation}
\setlength{\abovedisplayskip}{5pt}
\setlength{\belowdisplayskip}{5pt}
    \mathcal{L}_{\text{contact}}
    = \frac{1}{|\mathcal{G}_k|}
      \sum_{i \in \mathcal{G}_k}
      \max\bigl(0,\,-\cos\varphi_i \bigr),
\end{equation}
where $\cos\varphi_i\!=\!\frac{\mathbf{d}_i^\top \mathbf{d}_{k}}
            {\|\mathbf{d}_i\|\,\|\mathbf{d}_{k}\|}$ is the cosine similarity.

Since rigid parts should exhibit coherent motion, we employ a \textit{velocity consistency loss}~\cite{li2025garf,lipman2024flow,liu2022flow} by defining per-Gaussian displacements $\Delta\boldsymbol{\mu}_i\!=\!\boldsymbol{\mu}_i^1\!-\!\boldsymbol{\mu}_i^0$, and penalizing intra-part variance
\begin{equation}
\setlength{\abovedisplayskip}{6pt}
\setlength{\belowdisplayskip}{6pt}
    \mathcal{L}_{\text{velocity}} = \sum_{k=1}^K \text{Var}\left( \left\{ \Delta\boldsymbol{\mu}_i \mid i \in \mathcal{G}_k \right\} \right).
\end{equation}
We additionally employ a \textit{vector-field alignment loss} to ensure that predicted part transformations remain consistent with observed motion across different joint states. Inspired by flow-based models ~\cite{li2025garf,lipman2024flow,liu2022flow}, we treat part articulation as an SE(3) vector field acting on canonical Gaussians. For each part transformation $T_k\!=\!(\mathbf{R}_k, \mathbf{t}_k) \in \mathrm{SE}(3)$, we enforce consistency between predicted and observed positions 
\begin{equation}
\setlength{\abovedisplayskip}{5pt}
\setlength{\belowdisplayskip}{5pt}
    \mathcal{L}_{\text{vector}} = \sum_{k=1}^K \sum_{i \in \mathcal{G}_k} \left\| \mathbf{R}_k \boldsymbol{\mu}_i^0 + \mathbf{t}_k - \boldsymbol{\mu}_i^1 \right\|^2.
\end{equation}

\noindent \textbf{Training.} The overall training objective of \modelnamenc{} integrates reconstruction fidelity, part regularization, articulation learning, and physical consistency regularization. 
The total loss is defined as
\begin{equation}
\setlength{\abovedisplayskip}{5pt}
\setlength{\belowdisplayskip}{5pt}
    \mathcal{L}_{\text{\modelnamenc{}}}\!=\!\mathcal{L}_{\text{render}}\!+\!\lambda_{\text{part}} \mathcal{L}_{\text{part}}\!+\!\lambda_{\text{art}} \mathcal{L}_{\text{art}}\!+\!\lambda_{\text{phys}} \mathcal{L}_{\text{phys}}, 
\end{equation}
where $\mathcal{L}_{\text{phys}}\!=\!\mathcal{L}_{\text{contact}}\!+\!\mathcal{L}_{\text{velocity}}\!+\!\mathcal{L}_{\text{vector}}$, 
$\mathcal{L}_{\text{render}}$ is the rendering loss in \cref{eq:render_loss}, and  $\lambda_{\text{part}}$, $\lambda_{\text{art}}$, $\lambda_{\text{phys}}$ are coefficients. 

\begin{table*}[t]
\centering
\setlength\extrarowheight{1pt}
\caption{\textbf{Quantitative results on \textsc{\textbf{Paris}}}. Lower ($\downarrow$) is better across all metrics. \colorbox{cayenne!30}{\phantom{\rule{1ex}{1ex}}} highlights best performing results. Pos Err is omitted for prismatic joint only objects (Table 4 parts). Objects with \textbf{*} are seen categories trained in Ditto. F indicates wrong motion predictions.
}
\label{tab:main_result}
\vspace{-0.3cm}

\resizebox{0.99\textwidth}{!}{
\begin{tabular}{c c c *{10}{c} c c}
\toprule
& \multirow{2}{*}{\textbf{Metric}} 
& \multirow{2}{*}{\textbf{Method}} 
& \multicolumn{10}{c}{\textbf{Simulation}} 
& \multicolumn{2}{c}{\textbf{Real}} \\
\cmidrule(lr){4-13} \cmidrule(lr){14-15}
& & 
& \textbf{Foldchair} & \textbf{Fridge} & \textbf{Laptop*} & \textbf{Oven*} 
& \textbf{Scissor} & \textbf{Stapler} & \textbf{USB} & \textbf{Washer} 
& \textbf{Blade} & \textbf{Storage*} 
& \textbf{Real-Fridge} & \textbf{Real-Storage} \\
\midrule

\multirow{15}{*}{\rotatebox{90}{\textbf{Motion}}}
& \multirow{5}{*}{\textbf{Ang Err}}
& Ditto   & 89.35 & 89.30 & 3.12 & 0.96 & 4.50 & 89.86 & 89.77 & 89.51 & 79.54 & 6.32 & 1.71 & 5.88 \\
& & PARIS &19.05 &7.87 &0.03 &9.21 &22.34 &8.89 &0.82 &22.18 &50.45 &0.03 &9.92 &77.83 \\
& & DTA   & $0.03_{\pm 0.00}$ & $0.09_{\pm 0.00}$ & $0.07_{\pm 0.00}$ & $0.22_{\pm 0.10}$ 
        & $0.10_{\pm 0.00}$ & $0.07_{\pm 0.00}$ & $0.11_{\pm 0.00}$ & $0.36_{\pm 0.10}$ 
        & $0.20_{\pm 0.10}$ & $0.09_{\pm 0.00}$ & $2.08_{\pm 0.00}$ & $13.64_{\pm 3.60}$ \\
& & ArtGS & \cellcolor{cayenne!30}$0.01_{\pm 0.00}$ & $0.03_{\pm 0.00}$ 
        & \cellcolor{cayenne!30}$0.01_{\pm 0.00}$ & \cellcolor{cayenne!30}$0.01_{\pm 0.00}$
        & $0.05_{\pm 0.00}$ & \cellcolor{cayenne!30}$0.01_{\pm 0.00}$ & $0.04_{\pm 0.00}$ 
        & $0.02_{\pm 0.00}$ & $0.03_{\pm 0.00}$ & \cellcolor{cayenne!30}$0.01_{\pm 0.00}$
        & $2.09_{\pm 0.00}$ & $3.47_{\pm 0.30}$ \\
& & \textbf{\modelname{} (Ours)}
        & \cellcolor{cayenne!30}$0.01_{\pm 0.00}$ & \cellcolor{cayenne!30}$0.01_{\pm 0.00}$ 
        & \cellcolor{cayenne!30}$0.01_{\pm 0.00}$ & \cellcolor{cayenne!30}$0.01_{\pm 0.00}$ 
        & \cellcolor{cayenne!30}$0.02_{\pm 0.00}$ & \cellcolor{cayenne!30}$0.01_{\pm 0.00}$ 
        & \cellcolor{cayenne!30}$0.01_{\pm 0.00}$ & \cellcolor{cayenne!30}$0.01_{\pm 0.00}$ 
        & \cellcolor{cayenne!30}$0.01_{\pm 0.00}$ & $0.02_{\pm 0.00}$ 
        & \cellcolor{cayenne!30}$0.03_{\pm 0.01}$ & \cellcolor{cayenne!30}$1.24_{\pm 0.04}$ \\
\cmidrule(lr){2-15}

& \multirow{5}{*}{\textbf{Pos Err}}
& Ditto  & 3.77 & 1.02 & 0.01 & 0.13 & 5.70 & 0.20 & 5.41 & 0.66 & - & - & 1.84 & - \\
& & PARIS   & 0.35 & 3.13 & 0.04 & 0.07 & 2.59 & 7.67 & 6.35 & 4.05 & - & - & 1.50 & - \\
& & DTA & $0.01_{\pm 0.00}$ & $0.01_{\pm 0.00}$ & $0.01_{\pm 0.00}$ & $0.01_{\pm 0.00}$ & $0.02_{\pm 0.00}$ & $0.02_{\pm 0.00}$ & $0.00_{\pm 0.00}$ & $0.05_{\pm 0.00}$ & - & - & $0.59_{\pm 0.00}$ & - \\
& & ArtGS & \cellcolor{cayenne!30}$0.00_{\pm 0.00}$ & \cellcolor{cayenne!30}$0.00_{\pm 0.00}$ & $0.01_{\pm 0.00}$ & \cellcolor{cayenne!30}$0.00_{\pm 0.00}$ 
        & \cellcolor{cayenne!30}$0.00_{\pm 0.00}$ & \cellcolor{cayenne!30}$0.01_{\pm 0.00}$ & \cellcolor{cayenne!30}$0.00_{\pm 0.00}$ 
        & \cellcolor{cayenne!30}$0.00_{\pm 0.00}$ & - & - & $0.47_{\pm 0.00}$ & - \\
& & \textbf{\modelname{} (Ours)}
        & \cellcolor{cayenne!30}$0.00_{\pm 0.00}$ & \cellcolor{cayenne!30}$0.00_{\pm 0.00}$ 
        & \cellcolor{cayenne!30}$0.00_{\pm 0.00}$ & \cellcolor{cayenne!30}$0.00_{\pm 0.00}$ 
        & \cellcolor{cayenne!30}$0.00_{\pm 0.00}$ & \cellcolor{cayenne!30}$0.01_{\pm 0.00}$ 
        & \cellcolor{cayenne!30}$0.00_{\pm 0.00}$ & \cellcolor{cayenne!30}$0.00_{\pm 0.00}$ 
        & - & - & \cellcolor{cayenne!30}$0.13_{\pm 0.00}$ & - \\
\cmidrule(lr){2-15}

& \multirow{5}{*}{\textbf{Motion Err}}
& Ditto & 99.36 & F & 5.18 & 2.09 & 19.28 & 56.61 & 80.60 & 55.72 & F & 0.09 & 8.43 & 0.38 \\
& & PARIS & 166.24 & 102.34 & 0.03 & 28.18 & 124.38 & 117.71 & 167.98 & 126.77 & 0.38 & 0.36 & 2.68 & 0.58 \\
& & DTA   & $0.10_{\pm 0.00}$ & $0.12_{\pm 0.00}$ & $0.11_{\pm 0.00}$ & $0.12_{\pm 0.00}$ & $0.37_{\pm 0.60}$ & $0.08_{\pm 0.00}$ & $0.15_{\pm 0.00}$ & $0.28_{\pm 0.10}$ 
        & \cellcolor{cayenne!30}$0.00_{\pm 0.00}$ & \cellcolor{cayenne!30}$0.00_{\pm 0.00}$ & $1.85_{\pm 0.00}$ & $0.14_{\pm 0.00}$ \\
& & ArtGS & $0.03_{\pm 0.00}$ & $0.04_{\pm 0.00}$ & $0.02_{\pm 0.00}$ & $0.02_{\pm 0.00}$ & $0.04_{\pm 0.00}$ & $0.01_{\pm 0.00}$ & $0.03_{\pm 0.00}$ & $0.03_{\pm 0.00}$
        & \cellcolor{cayenne!30}$0.00_{\pm 0.00}$ & \cellcolor{cayenne!30}$0.00_{\pm 0.00}$ & $1.94_{\pm 0.00}$ & $0.04_{\pm 0.00}$ \\
& & \textbf{\modelname{} (Ours)} 
        & \cellcolor{cayenne!30}$0.01_{\pm 0.00}$ & \cellcolor{cayenne!30}$0.01_{\pm 0.00}$ & \cellcolor{cayenne!30}$0.01_{\pm 0.00}$ & \cellcolor{cayenne!30}$0.00_{\pm 0.00}$ 
        & \cellcolor{cayenne!30}$0.01_{\pm 0.00}$ & \cellcolor{cayenne!30}$0.00_{\pm 0.00}$ & \cellcolor{cayenne!30}$0.01_{\pm 0.00}$ & \cellcolor{cayenne!30}$0.02_{\pm 0.00}$ 
        & \cellcolor{cayenne!30}$0.00_{\pm 0.00}$ & \cellcolor{cayenne!30}$0.00_{\pm 0.00}$ & \cellcolor{cayenne!30}$0.72_{\pm 0.01}$ & \cellcolor{cayenne!30}$0.02_{\pm 0.01}$ \\
\midrule

\multirow{15}{*}{\rotatebox{90}{\textbf{Geometry}}}
& \multirow{5}{*}{\textbf{CD$_{\textsubscript{static}}$}}
& Ditto & 33.79 & 3.05 & 0.25 & \cellcolor{cayenne!30}2.52 & 39.07 & 41.64 & 2.64 & 10.32 & 46.90 & 9.18 & 47.01 & 16.09 \\
& & PARIS & 11.21 & 11.78 & 0.17 & 3.58 & 17.88 & 4.79 & 2.41 & 15.92 & 2.24 & 9.83 & 13.79 & 23.92 \\
& & DTA & $0.18_{\pm 0.00}$ & $0.62_{\pm 0.00}$ & $0.30_{\pm 0.00}$ & $4.60_{\pm 0.10}$ & $3.55_{\pm 6.10}$ & $2.91_{\pm 0.10}$ & $2.32_{\pm 0.10}$ & $4.56_{\pm 0.10}$ & $0.55_{\pm 0.00}$ & $4.90_{\pm 0.50}$ & $2.36_{\pm 0.10}$ & $10.98_{\pm 0.10}$ \\
& & ArtGS & $0.26_{\pm 0.30}$ & $0.52_{\pm 0.00}$ & $0.63_{\pm 0.00}$ & $3.88_{\pm 0.00}$ & $0.61_{\pm 0.30}$ & $3.83_{\pm 0.10}$ & $2.25_{\pm 0.20}$ & $6.43_{\pm 0.10}$ & $0.54_{\pm 0.00}$ & $7.31_{\pm 0.20}$ & $1.64_{\pm 0.20}$ & $2.93_{\pm 0.30}$ \\
& & \textbf{\modelname{} (Ours)}
        & \cellcolor{cayenne!30}$0.14_{\pm 0.00}$ & \cellcolor{cayenne!30}$0.41_{\pm 0.00}$ & \cellcolor{cayenne!30}$0.15_{\pm 0.00}$ 
        & $2.91_{\pm 0.01}$ & \cellcolor{cayenne!30}$0.48_{\pm 0.01}$ & \cellcolor{cayenne!30}$2.36_{\pm 0.03}$ & \cellcolor{cayenne!30}$1.84_{\pm 0.03}$ 
        & \cellcolor{cayenne!30}$3.92_{\pm 0.02}$ & \cellcolor{cayenne!30}$0.42_{\pm 0.00}$ & \cellcolor{cayenne!30}$3.58_{\pm 0.00}$ 
        & \cellcolor{cayenne!30}$1.29_{\pm 0.01}$ & \cellcolor{cayenne!30}$2.12_{\pm 0.02}$\\
\cmidrule(lr){2-15}
& \multirow{5}{*}{\textbf{CD$_{\textsubscript{movable}}$}}
& Ditto & 141.11 & 0.99 & 0.19 & 0.94 & 20.68 & 31.21 & 15.88 & 12.89 & 195.93 & 2.20 & 50.60 & 20.35 \\
& & PARIS & 24.23 &12.88 &0.17 &7.49 &18.89 &38.42 &13.81 &379.40 &200.24 &63.97 &91.72 &528.83 \\
& & DTA & $0.15_{\pm 0.00}$ & $0.27_{\pm 0.00}$ & $0.13_{\pm 0.00}$ & $0.44_{\pm 0.00}$ & $10.11_{\pm 19.40}$ & $1.13_{\pm 0.50}$ & \cellcolor{cayenne!30}$1.47_{\pm 0.00}$ & $0.45_{\pm 0.00}$ 
        & $2.05_{\pm 0.30}$ & \cellcolor{cayenne!30}$0.36_{\pm 0.00}$ & $1.12_{\pm 0.00}$ & $30.78_{\pm 2.60}$ \\
& & ArtGS & $0.54_{\pm 0.10}$ & $0.21_{\pm 0.00}$ & $0.13_{\pm 0.00}$ & $0.89_{\pm 0.20}$ & $0.64_{\pm 0.40}$ & $0.52_{\pm 0.10}$ & $1.22_{\pm 0.10}$ & $0.45_{\pm 0.20}$ 
        & \cellcolor{cayenne!30}$1.12_{\pm 0.20}$ & $1.02_{\pm 0.40}$ & $0.66_{\pm 0.20}$ & $6.28_{\pm 3.60}$ \\
& & \textbf{\modelname{} (Ours)}
        & \cellcolor{cayenne!30}$0.12_{\pm 0.00}$ & \cellcolor{cayenne!30}$0.18_{\pm 0.01}$ & \cellcolor{cayenne!30}$0.11_{\pm 0.00}$ & \cellcolor{cayenne!30}$0.38_{\pm 0.00}$ 
        & \cellcolor{cayenne!30}$0.51_{\pm 0.01}$ & \cellcolor{cayenne!30}$0.41_{\pm 0.00}$ & $1.05_{\pm 0.00}$ & \cellcolor{cayenne!30}$0.39_{\pm 0.00}$ 
        & $1.42_{\pm 0.01}$ & $0.78_{\pm0.00}$ & \cellcolor{cayenne!30}$0.55_{\pm 0.01}$ & \cellcolor{cayenne!30}$5.01_{\pm 0.03}$ \\
\cmidrule(lr){2-15}

& \multirow{5}{*}{\textbf{CD$_{\textsubscript{whole}}$}}
& Ditto & 6.80 & 2.16 & 0.31 & 2.51 & 1.70 & 2.38 & 2.09 & 7.29 & 42.04 & 3.91 & 6.50 & 14.08 \\
& & PARIS & 8.22 & 9.31 & 0.28 & 5.44 & 6.13 & 9.62 & 2.14 & 14.35 & 0.76 & 9.62 & 11.52 & 38.94 \\
& & DTA & $0.27_{\pm 0.00}$ & $0.70_{\pm 0.00}$ & $0.32_{\pm 0.00}$ & $4.24_{\pm 0.01}$ & \cellcolor{cayenne!30}$0.41_{\pm 0.00}$ & $1.92_{\pm 0.00}$ & $1.17_{\pm 0.00}$ & $4.48_{\pm 0.20}$ & $0.36_{\pm 0.00}$ & $3.99_{\pm 0.40}$ & $2.08_{\pm 0.10}$ & $8.98_{\pm 0.10}$ \\
& & ArtGS & $0.43_{\pm 0.20}$ & $0.58_{\pm 0.00}$ & $0.50_{\pm 0.00}$ & $3.58_{\pm 0.00}$ & $0.67_{\pm 0.30}$ & $2.63_{\pm 0.00}$ & $1.28_{\pm 0.00}$ & $5.99_{\pm 0.10}$ & $0.61_{\pm 0.00}$ & $5.21_{\pm 0.10}$ & $1.29_{\pm 0.10}$ & $3.23_{\pm 0.10}$ \\
& & \textbf{\modelname{} (Ours)}
        & \cellcolor{cayenne!30}$0.19_{\pm 0.00}$ & \cellcolor{cayenne!30}$0.43_{\pm 0.00}$ & \cellcolor{cayenne!30}$0.20_{\pm 0.00}$ 
        & \cellcolor{cayenne!30}$1.85_{\pm 0.01}$ & $0.42_{\pm 0.00}$ & \cellcolor{cayenne!30}$1.45_{\pm 0.01}$ 
        & \cellcolor{cayenne!30}$0.92_{\pm 0.01}$ & \cellcolor{cayenne!30}$3.45_{\pm 0.02}$ 
        & \cellcolor{cayenne!30}$0.35_{\pm 0.00}$ & \cellcolor{cayenne!30}$2.87_{\pm 0.01}$ 
        & \cellcolor{cayenne!30}$1.03_{\pm 0.00}$ & \cellcolor{cayenne!30}$2.78_{\pm 0.01}$ \\
\bottomrule
\end{tabular}}
\end{table*}

\section{Experiments}
\label{sec:experiments}
We compare \modelnamenc{} against Ditto \cite{jiang2022ditto}, PARIS \cite{liu2023paris}, ArtGS \cite{liu2025building}, and DTA \cite{weng2024neural} on three object articulation datasets with varying levels of articulation complexity: \textsc{Paris} \cite{liu2023paris} (10 synthetic objects with 1 movable part),  \textsc{ArtGS-Multi} \cite{liu2025building} (5 synthetic objects with 3–6 movable parts), and \textsc{DTA-Multi} \cite{weng2024neural} (2 synthetic objects with 2 movable parts).

Following prior articulated object modeling work~\cite{jiang2022ditto, liu2023paris, liu2025building}, to assess geometry quality, we report Chamfer Distance scores separately for the entire object ($\text{CD}_{\text{whole}}$), the static components ($\text{CD}_{\text{static}}$), and the average of the movable parts ($\text{CD}_{\text{movable}}$). To assess articulation accuracy, we measure the angular deviation between the predicted and actual joint axes (Ang Err), the positional offset for revolute joints (Pos Err), and the part motion error (Motion Err). 
Additional implementation details can be found in Appendix \ref{app:impl}. 

\subsection{Experimental Results}
Table~\ref{tab:main_result} reports results on the \textsc{Paris} benchmark. \modelnamenc{} achieves the lowest errors across all metrics, accurately recovering joint parameters and articulations. The average angular error remains below $0.01^\circ$ on nearly all simulated objects, over two orders of magnitude lower than Ditto~\citep{jiang2022ditto} and PARIS~\citep{liu2023paris}. 
For revolute joints, \modelnamenc{} achieves near-zero positional error, indicating highly accurate recovery of motion axes.
On motion accuracy, measured by geodesic or Euclidean distance depending on joint type, \modelnamenc{} also leads with near-zero error on most categories. This highlights the benefit of our motion-consistent design.

In terms of geometry, \modelnamenc{} consistently achieves higher geometric fidelity, reducing Chamfer Distance across all categories by up to 1.74$\times$ relative to the next best baseline, while delivering a 2-4$\times$ improvement over DTA and ArtGS on both static and dynamic geometry. 
In contrast to ArtGS, which relies on heuristic Gaussian clustering, \modelnamenc{} learns soft part-identity embeddings jointly with physics-guided constraints, enabling coherent part boundaries to emerge directly from spatial and kinematic cues. As a result, \modelnamenc{} attains consistently lower $\text{CD}_{\text{movable}}$ and $\text{CD}_{\text{whole}}$, indicating more accurate and stable reconstruction of articulated parts. The learned representation also eliminates part drift, as indicated by the near-zero $\text{MotionErr}$, and more effectively suppresses interpenetration, yielding a 4–10$\times$ reduction in the most challenging metric $\text{CD}_{\text{movable}}$ compared to ArtGS. Collectively, these gains lead to sharper part segmentation and more physically consistent articulation.

\begin{table*}[t]
    \centering
    \setlength\extrarowheight{1pt}
    \caption{\textbf{Quantitative results on \textsc{\textbf{\textbf{DTA}}-Multi} and \textsc{\textbf{\textbf{ArtGS}}-Multi}}. Lower ($\downarrow$) is better across all metrics. \colorbox{cayenne!30}{\phantom{\rule{1ex}{1ex}}} highlights best performing results. Pos Err is omitted for prismatic-only objects (Table 4 parts).
    }
    \vspace{-0.3cm}
    \label{tab:complex_main_result}
    \resizebox{0.99\textwidth}{!}{%
    \begin{tabular}{c c  >{\centering\arraybackslash}p{3.5cm}   c c c c c c c c c c  c c}
        \toprule
        \multirow{2}{*}{\textbf{Category}} & \multirow{2}{*}{\textbf{Metric}} & \multirow{2}{*}{\textbf{Method}} & \textbf{{\begin{tabular}{c} \textbf{Fridge} \\ \textbf{(3 parts)} \end{tabular}}} & \textbf{{\begin{tabular}{c} \textbf{Table} \\ \textbf{(4 parts)} \end{tabular}}} &{\begin{tabular}{c} \textbf{Table} \\ \textbf{(5 parts)} \end{tabular}} &{\begin{tabular}{c} \textbf{Storage} \\ \textbf{(3 parts)} \end{tabular}} & {\begin{tabular}{c} \textbf{Storage} \\ \textbf{(4 parts)} \end{tabular}} & {\begin{tabular}{c} \textbf{Storage} \\ \textbf{(7 parts)} \end{tabular}}  & {\begin{tabular}{c} \textbf{Oven} \\ \textbf{(4 parts)} \end{tabular}} \\
        \midrule
        
        \multirow{9}{*}{\rotatebox{90}{\textbf{Motion}}} & \multirow{3}{*}{\shortstack{\textbf{Ang}\\\textbf{Err}}} 
        & \textbf{DTA} & 0.16 & 24.35 & 20.62 & 0.29 & 51.18 & 19.07 & 17.83 \\
        & & \textbf{ArtGS} & \cellcolor{cayenne!30}0.01 & 1.16 & 0.04 & 0.02 & 0.02 & 0.14 & 0.04 \\
        & & \textbf{\modelname{} (Ours)} & \cellcolor{cayenne!30}0.01 & \cellcolor{cayenne!30}0.08 & \cellcolor{cayenne!30}0.03 & \cellcolor{cayenne!30}0.01 & \cellcolor{cayenne!30}0.01 & \cellcolor{cayenne!30}0.11 & \cellcolor{cayenne!30}0.03 \\

        \cmidrule(lr){2-10}
        
        & \multirow{3}{*}{\shortstack{\textbf{Pos}\\\textbf{Err}}} 
        & \textbf{DTA} & 0.01 & - & 4.2 & 0.04 & 2.44 & 0.31 & 6.51 \\
        & & \textbf{ArtGS} & \cellcolor{cayenne!30}0.00 & - & \cellcolor{cayenne!30}0.00 & 0.01 & \cellcolor{cayenne!30}0.00 & 0.02 & \cellcolor{cayenne!30}0.01 \\
        & & \textbf{\modelname{} (Ours)} & \cellcolor{cayenne!30}0.00 & - & \cellcolor{cayenne!30}0.00 & \cellcolor{cayenne!30}0.00 & \cellcolor{cayenne!30}0.00 & \cellcolor{cayenne!30}0.01 & \cellcolor{cayenne!30}0.01 \\

        \cmidrule(lr){2-10}

        & \multirow{3}{*}{\shortstack{\textbf{Motion}\\\textbf{Err}}} 
        & \textbf{DTA} & 0.16 & 0.12 & 30.8 & 0.07 & 43.77 & 10.67 & 31.80 \\
        & & \textbf{ArtGS} & 0.03 & \cellcolor{cayenne!30}0.00 & \cellcolor{cayenne!30}0.01 & \cellcolor{cayenne!30}0.01 & 0.03 & 0.62 & 0.23 \\
        & & \textbf{\modelname{} (Ours)} & \cellcolor{cayenne!30}0.02 & \cellcolor{cayenne!30}0.00 & \cellcolor{cayenne!30}0.01 & \cellcolor{cayenne!30}0.01 & \cellcolor{cayenne!30}0.02 & \cellcolor{cayenne!30}0.55 & \cellcolor{cayenne!30}0.18 \\

        \midrule
        \multirow{9}{*}{\rotatebox{90}{\textbf{Geometry}}} & \multirow{3}{*}{\textbf{CD\textsubscript{static}}} 
        & \textbf{DTA} & 0.63 & 0.59 & 1.39 & 0.86 & 5.74 & 0.82 & 1.17 \\
        & & \textbf{ArtGS} & 0.62 & 0.74 & 1.22 & 0.78 & 0.75 & 0.67 & 1.08 \\
        & & \textbf{\modelname{} (Ours)} & \cellcolor{cayenne!30}0.59 & \cellcolor{cayenne!30}0.56 & \cellcolor{cayenne!30}1.18 & \cellcolor{cayenne!30}0.73 & \cellcolor{cayenne!30}0.68 & \cellcolor{cayenne!30}0.61 & \cellcolor{cayenne!30}1.01 \\

        \cmidrule(lr){2-10}
        
        & \multirow{3}{*}{\textbf{CD\textsubscript{movable}}} 
        & \textbf{DTA} & 0.48 & 104.38 & 230.38 & 0.23 & 246.63 & 476.91 & 359.16 \\
        & & \textbf{ArtGS} & 0.13 & 3.53 & 3.09 & 0.23 & 0.13 & 3.70 & 0.25 \\
        & & \textbf{\modelname{} (Ours)} & \cellcolor{cayenne!30}0.08 & \cellcolor{cayenne!30}1.95 & \cellcolor{cayenne!30}1.85 & \cellcolor{cayenne!30}0.09 & \cellcolor{cayenne!30}0.07 & \cellcolor{cayenne!30}1.83 & \cellcolor{cayenne!30}0.11 \\

        \cmidrule(lr){2-10}

        & \multirow{3}{*}{\textbf{CD\textsubscript{whole}}} 
        & \textbf{DTA} & 0.88 & 0.55 & \cellcolor{cayenne!30}1.00 & 0.97 & 0.88 & 0.71 & 1.01 \\
        & & \textbf{ArtGS} & 0.75 & 0.74 & 1.16 & 0.93 & 0.88 & 0.70 & 1.03 \\
        & & \textbf{\modelname{} (Ours)} & \cellcolor{cayenne!30}0.73 & \cellcolor{cayenne!30}0.51 & 1.10 & \cellcolor{cayenne!30}0.87 & \cellcolor{cayenne!30}0.80 & \cellcolor{cayenne!30}0.63 & \cellcolor{cayenne!30}0.95 \\
        \bottomrule
    \end{tabular}
    }
\end{table*}
\begin{table*}[t!]
    \centering
    \scriptsize
    \setlength{\tabcolsep}{2pt}
    \caption{\textbf{\modelname{} key component ablations} on the two most complex objects in our evaluation, {Table (5 parts)} and {Storage (7 parts)}. Lower ($\downarrow$) is better on all metrics. We add each component cumulatively, starting from vanilla.  
    \colorbox{cayenne!30}{\phantom{\rule{1ex}{1ex}}} highlights the best results.}
    \label{tab:main_ablation_v2}
    \vspace{-0.3cm}
    \begin{tabularx}{0.99\textwidth}{c c *{6}{>{\centering\arraybackslash}X}}
        \toprule
        \textbf{Objects} & \textbf{Methods} &
        \textbf{AngErr} & \textbf{PosErr} & \textbf{MotionErr} &
        \textbf{CD\textsubscript{static}} &  \textbf{CD\textsubscript{movable}} &  \textbf{CD\textsubscript{whole}} \\
        \cmidrule{1-8}
        \multirow{4}{*}{{\begin{tabular}{c}\textbf{Table}\\\textbf{(5 parts)}\end{tabular}}}
            & \textbf{Vanilla} &  17.32 &  1.01 &  27.64 &  7.11 &  132.21 &  2.78 \\
            & +~\textbf{part parameters} & 0.28 & 0.19 & 2.35 & 2.65 & 28.35 & 1.52 \\
            & +~\textbf{repel points} & 0.05 & 0.03 & 0.18 & 1.32 & 4.47 & 1.65 \\
            & +~\textbf{physical constraints} (\modelname{})                      & \cellcolor{cayenne!30} 0.03 & \cellcolor{cayenne!30} 0.00 & \cellcolor{cayenne!30} 0.01 & \cellcolor{cayenne!30} 1.18 & \cellcolor{cayenne!30} 1.85 & \cellcolor{cayenne!30} 1.10 \\
        \hline
        \multirow{4}{*}{{\begin{tabular}{c}\textbf{Storage}\\\textbf{(7 parts)}\end{tabular}}}
            & \textbf{Vanilla} &  27.24 &  1.32 &  24.41 &  11.23 &  497.17 &  2.74 \\
            & +~\textbf{part parameters} & 0.91 & 0.28 & 2.61 & 4.02 & 15.68 & 1.89 \\
            & +~\textbf{repel points} & 0.14 & 0.05 & 0.04 & 1.22 & 4.54 & 1.12 \\
            & +~\textbf{physical constraints} (\modelname{})  & 
            \cellcolor{cayenne!30} 0.11 & \cellcolor{cayenne!30} 0.01 & \cellcolor{cayenne!30} 0.55 & \cellcolor{cayenne!30} 0.61 & \cellcolor{cayenne!30} 1.83 & \cellcolor{cayenne!30} 0.63 \\
        \bottomrule
    \end{tabularx}
\end{table*}

Table~\ref{tab:complex_main_result} presents results on the \textsc{{DTA}-Multi} and \textsc{{ArtGS}-Multi} benchmarks, which contain objects with multiple movable parts. \modelnamenc{} consistently outperforms DTA and ArtGS across all objects and metrics. In terms of articulation accuracy, \modelnamenc{} achieves the lowest angular and positional errors on nearly every example, with particularly strong gains in motion error, where \modelnamenc{} matches or surpasses the strongest baseline (ArtGS) even on challenging multi-part objects such as \textsc{Storage} (7 articulated parts).

In terms of geometry, \modelnamenc{} attains the lowest Chamfer Distance for static, movable, and whole-object regions in almost all categories. The largest improvements appear in $\text{CD}_{\text{movable}}$, where the proposed part-aware representation reduces error by up to 10$\times$ over DTA and 3$\times$ over ArtGS. This confirms that the learned parts enable robust part discovery and articulation, whereas competing methods often exhibit part drift or under-segmentation.

Moreover, we assess statistical significance using t-tests (n = 3) for each object-metric pair comparing \modelnamenc{} against ArtGS. To keep the analysis conservative and avoid overstating improvements, we use a small epsilon (1e-6). Across all 111 object-metric pairs evaluated, \modelnamenc{} achieves statistically significant  ($p<0.05$) over ArtGS in 83 cases, shows no statistically significant difference in 25 cases, and performs worse in only 3, confirming the consistency and reliability of the gains obtained by \modelnamenc{}.

\subsection{Ablations}
We conduct ablations to evaluate the contribution of three key \modelnamenc{} components: part ID parameters, repulsion points, and physical constraints. We select two of the most complex objects, Table (5 parts) and Storage (7 parts), to examine performance under challenging settings. As shown in~\Cref{tab:main_ablation_v2}, each component progressively improves both articulation and geometry accuracy.

\noindent \textbf{Part Parameters.} Introducing part parameters yields the most significant improvement across all metrics. 
For the 5-part Table, angular error drops from 17.32$\rightarrow$0.28 and motion error from 27.64$\rightarrow$2.35, a {$>$90\% reduction} in both, while $\text{CD}_{\text{movable}}$ decreases from 132.21$\rightarrow$28.35, showing {$\sim$4.6$\times$ improvement} in geometric fidelity.
On the most complex 7-part Storage object, angular error decreases from 27.24$\rightarrow$0.91 and motion error from 24.41$\rightarrow$2.61, a nearly {10$\times$ improvement}, while $\text{CD}_{\text{movable}}$ drops from 497.17$\rightarrow$15.68, representing a {$\sim$32$\times$ reduction} in geometric error. These results demonstrate that accurate part segmentation is foundational for both geometry and articulation, allowing the model to disentangle and track rigid parts effectively.\looseness-1

\noindent \textbf{Repel Points.} Incorporating repel points further enhances motion quality by enforcing inter-part separation. 
On 5-part Table, motion error {drops by $\sim$92\%} (2.35$\rightarrow$0.18) and $\text{CD}_{\text{movable}}$ {drops by $\sim$84\%} (28.35$\rightarrow$4.47). 
For 7-part Storage, motion error drops by $\sim${98\%} (2.61$\rightarrow$0.04) and $\text{CD}_{\text{movable}}$ by {70\%} (15.68$\rightarrow$4.54). 
These improvements confirm that spatial repulsion effectively prevents interpenetration.

\noindent \textbf{Physical Constraints.} 
Finally, introducing {physical constraints} yields the best overall performance across all metrics.
On the 5-part Table, motion error is reduced by another {$\sim$94\%} (0.18$\rightarrow$0.01), while $\text{CD}_{\text{movable}}$ decreases from 4.47$\rightarrow$1.85. On the 7-part Storage, $\text{CD}_{\text{movable}}$ further decreases from 4.54$\rightarrow$1.83, while preserving low motion errors. 
Physical constraints act as effective regularizers to enforce physical plausibility by encouraging consistent part trajectories, preserving joint-compatible motion, and preventing collisions across articulated states.
In summary, our part-aware design is most crucial for capturing semantic structure, while repulsion and physical priors further enhance geometric accuracy and articulation quality.

\begin{figure}[t]
\centering
\includegraphics[width=0.99\linewidth]{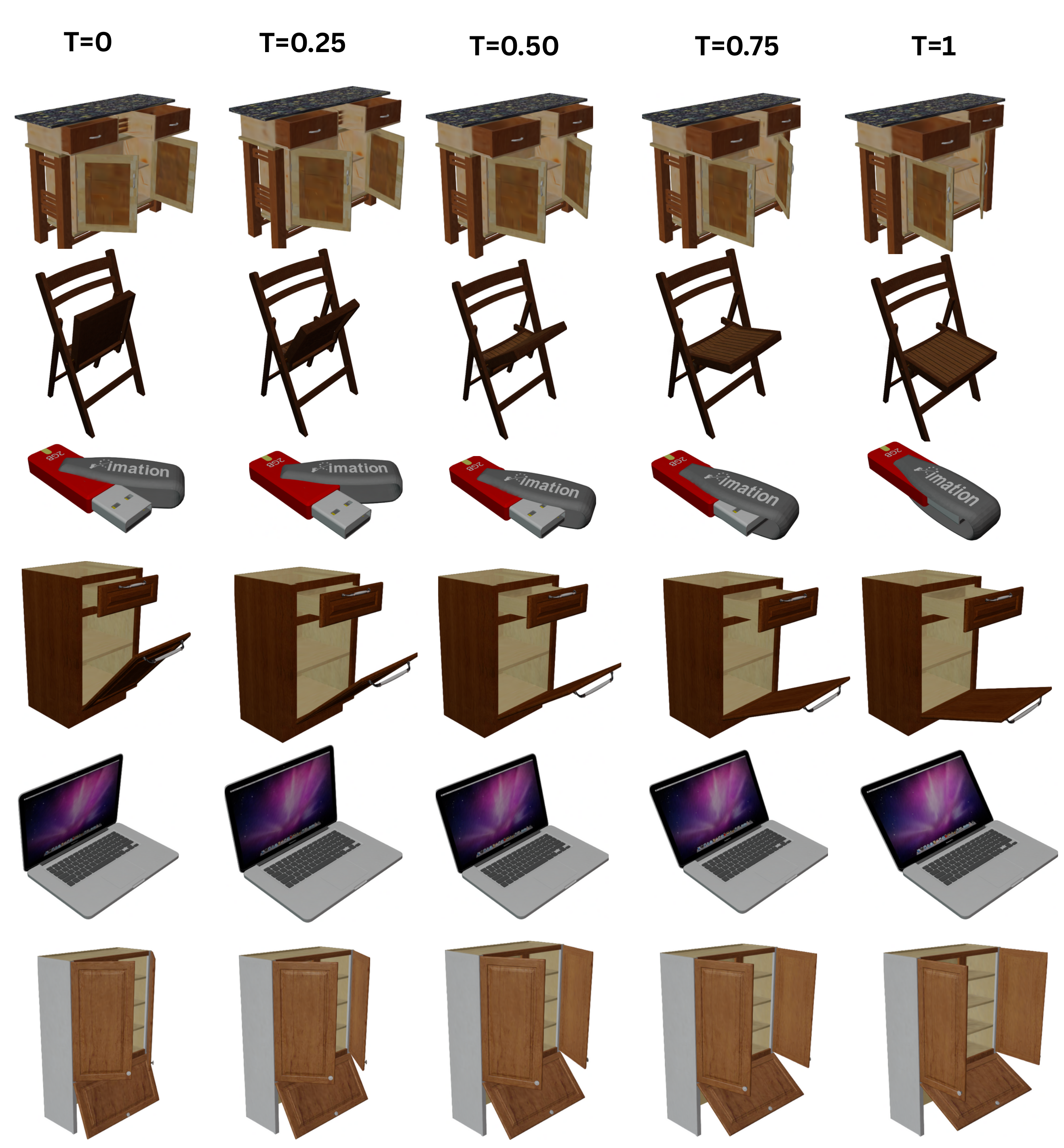}
\vspace{-0.2cm}
\caption{\textbf{\modelname{} Qualitative examples of articulated assets} across six objects consisting of both single part (USB, Foldchair, Laptop) and multi-part (Table, Storage, Cupboard) articulations.}
\label{fig:results_grid_full}
\end{figure}

\subsection{Qualitative Results} \Cref{fig:results_grid_full} presents qualitative articulation results across six articulated objects with varied joint types and geometries, demonstrating that \modelnamenc{} produces smooth, physically plausible motion trajectories from the fully closed state (T = 0) to the fully open state (T = 1). 
Each row shows a different object undergoing continuous motion, with smooth transitions between configurations. These intermediate frames demonstrate that \modelnamenc{} produces consistent motion paths through the full articulation sequence, highlighting our model’s ability to produce realistic motions and generalize across both single-part and complex multi-part articulations.

\begin{figure}[t!]
  \centering
  \includegraphics[width=0.99\linewidth]{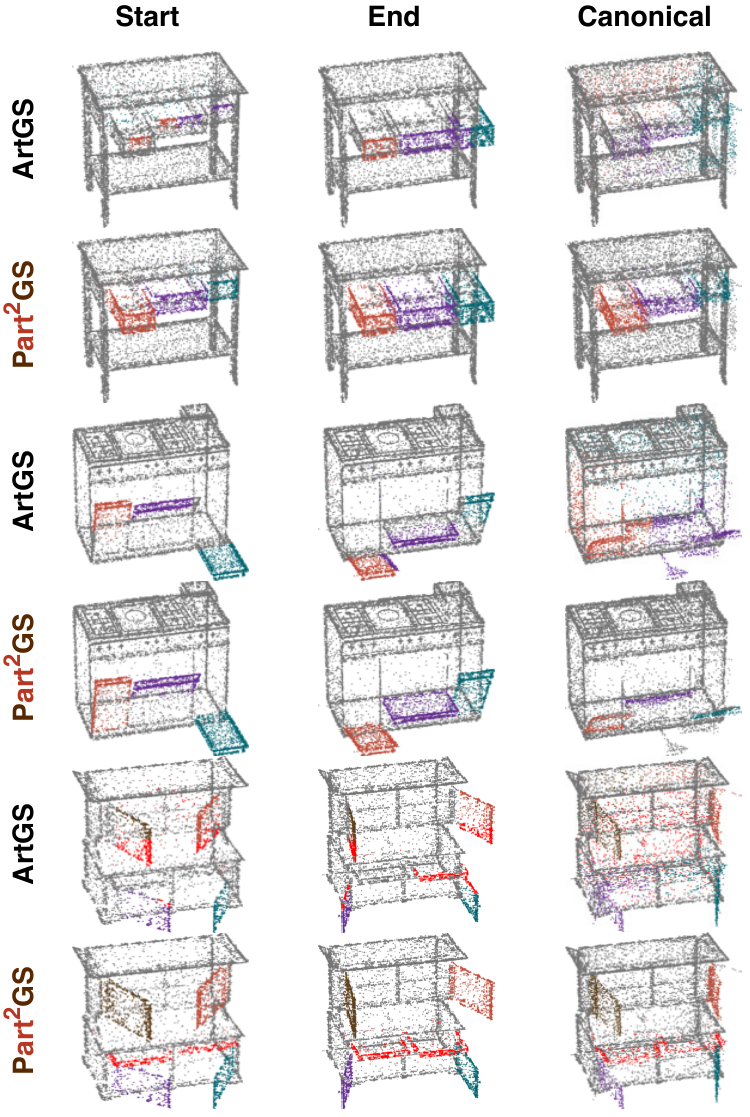}
  \vspace{-0.2cm}
  \caption{\textbf{Qualitative comparison of part discovery across object states (columns)}. 
  \modelname{} accurately isolates moving parts, whereas ArtGS struggles to maintain distinct part groupings, leading to blurred or collapsed representations.}
  \label{fig:part_qualitativebig}
  \vspace{-0.2cm}
\end{figure}

\Cref{fig:part_qualitativebig} shows a qualitative comparison of the part assignments produced by \modelnamenc{} and ArtGS in their canonical representations. Examples show \modelnamenc{} produces clean, consistent segmentation across all configurations. In both start and end states, \modelnamenc{} accurately isolates moving parts (\eg drawers and doors) with minimal leakage. In the canonical state, our method retains sharp part boundaries, demonstrating robust part identification under challenging intermediate configurations. This indicates that encoding motion information into the canonical Gaussian initialization is critical for obtaining a clean, part-aware canonical space that downstream articulation optimization can reliably refine.\looseness-1

\section{Conclusion}
We introduce \modelname{}, a part-aware framework for reconstructing articulated 3D digital twins directly from raw multi-view observations. By coupling learnable part-aware Gaussian representations with motion-aware canonicalization, physics-guided regularization, and repel-point-based articulation refinement, \modelnamenc{} recovers articulated structure, high-fidelity geometry, and physically coherent motion within a unified 3D Gaussian Splatting formulation. Unlike prior approaches that rely on heuristic clustering, direct pose interpolation, or external structural priors, the proposed framework enables part boundaries and articulation behavior to emerge jointly from geometric, kinematic, and physical cues. Extensive experiments across diverse articulation settings show that \modelnamenc{} consistently improves reconstruction quality and articulation accuracy, including substantial gains on challenging multi-part settings.

\section*{Acknowledgments}
This research was partially supported by Google, the Google TPU Research Cloud (TRC) program, the U.S. Defense Advanced Research Projects Agency (DARPA) under award HR001125C0303, and the U.S. Army under contract W5170125CA160. The views and conclusions contained herein are those of the authors and should not be interpreted as necessarily representing the official policies, either expressed or implied, of Google, DARPA, the U.S. Army, or the U.S. Government. The U.S. Government is authorized to reproduce and distribute reprints for governmental purposes notwithstanding any copyright annotation therein.\\

{
    \small
    \bibliographystyle{ieeenat_fullname}
    \bibliography{main}
}

\appendix
\clearpage
\setcounter{page}{1}
\maketitlesupplementary

\section{Implementation Details}\label{app:impl}
\noindent\textbf{Part Assignment Details.}
As defined in \Cref{sec:partdisc}, the part identity of a Gaussian $G_i$ is represented by a continuous probability distribution $F(G_i)=\text{softmax}(f(\boldsymbol{\psi}_i))$. To maintain full differentiability, we employ a soft, probability-weighted strategy for applying transformations. 

The final transformed position $\boldsymbol{\mu}_i^{(t)}$ of Gaussian $G_i$ is computed as a weighted sum over all $K$ possible part transformations $\mathcal{T}=\{T_k\}_{k=1}^K$:
\begin{equation}
\setlength{\abovedisplayskip}{6pt}
\setlength{\belowdisplayskip}{6pt}
\boldsymbol{\mu}_i^{(t)}
=
\sum_{k=1}^K p_{i,k}\,\big(\mathbf{R}_k^{(t)} \boldsymbol{\mu}_i^0 + \mathbf{t}_k^{(t)}\big)
+ \mathbf{F}_{\text{repel}, i}.
\end{equation}
Here, $p_{i,k}$ denotes the probability that Gaussian $G_i$ belongs to part $k$. This formulation enables the articulation and consistency losses to jointly optimize both the part-identity embedding $\boldsymbol{\psi}_i$ and the transformation parameters $(\mathbf{R}_k,\mathbf{t}_k)$. During inference, each Gaussian is assigned the rigid transformation of its most likely part, given by $k^*=\operatorname*{argmax}_k F(G_i)$.

\noindent\textbf{Part Supervision.} Our method does not require explicit part-level supervision, but it does assume a user-specified upper bound on the number of possible part groups, denoted by \(K\). Specifying \(K\) does not introduce supervision for the following reasons: (1) The model is never told which part corresponds to which semantic region; it must infer part clusters entirely through geometric and motion consistency losses. (2) The KL-based neighborhood regularization (\Cref{sec:partdisc}) forces part probabilities to self-organize based purely on geometric affinity. Thus, the method remains fully self-supervised with respect to part identity. 

We also analyze the effect of misspecifying the number of part \(K\). ~\Cref{tab:k_ablation} shows that under-specifying $K$ significantly degrades accuracy, while over-specifying it causes only mild degradation. Under-specifying $K$ forces multiple physically distinct parts to share a single rigid slot. Because each slot models only one SE(3) motion, merging parts with different joint axes produces inconsistent transformations, leading to large errors in motion estimation and geometry reconstruction.
In contrast, over-specifying $K$ introduces extra slots that receive no coherent geometric or kinematic signal. These redundant slots naturally collapse due to the part regularizer, velocity-consistency loss, and articulation constraints, resulting in only mild degradation.

\noindent \textbf{Repel Point Initialization.}
In our formulation, repel points are placed only on the static base and used to discourage interpenetration from movable parts. We perform an ablation on the most complex object Storage (7 parts), adopting a slightly more general and stable strategy. Specifically, we first use the canonical Gaussians to identify locations where movable parts lie within a small distance threshold of the static base. We then uniformly sample $N_R\!=\!2000$ repel points from these proximity regions, which naturally concentrates repulsion forces along potential contact interfaces. These repel points remain fixed throughout training and are not updated or pruned, preventing drift and keeping the optimization stable.

As shown in ~\Cref{tab:repel_num_ablation}, performance remains stable across all tested values, with no noticeable impact on final articulation accuracy. Using too few repel points slightly increases transient overlap at early iterations, but it does not affect convergence. Increasing $N_R$ provides no measurable benefit, confirming that our method does not depend on problem-specific tuning. Because repel points act as a soft collision prior and are not tied to any assumptions about joint type or motion, the model naturally corrects for noisy or imperfect repel placement during optimization.

\begin{table}[t]
    \centering
    \caption{\textbf{Specifying \# parts.}  
    Lower ($\downarrow$) is better across all metrics. 
    \colorbox{cayenne!30}{\phantom{\rule{1ex}{1ex}}} highlights the best-performing setting.}
    \vspace{-0.3cm}
    \label{tab:k_ablation}
    \resizebox{0.99\linewidth}{!}{
    \begin{tabular}{c c c c c | c  c c c}
    \toprule
    \textbf{K} & \textbf{Metric}  &
    \shortstack{\textbf{Storage}\\\textbf{(4 parts)}} &
    \shortstack{\textbf{Oven}\\\textbf{(4 parts)}} &
    \shortstack{\textbf{Table}\\\textbf{(4 parts)}} &
    \textbf{Metric} &
    \shortstack{\textbf{Storage}\\\textbf{(4 parts)}} &
    \shortstack{\textbf{Oven}\\\textbf{(4 parts)}} &
    \shortstack{\textbf{Table}\\\textbf{(4 parts)}} \\
    \hline

     2 & \multirow{5}{*}{\textbf{Ang Err}}   
    &  0.12 & 0.20 & 0.25 
         & \multirow{5}{*}{\textbf{CD$_{\text{static}}$}}
         & 4.90 & 2.30 & 14.80 \\
     3 &  & 0.06 & 0.12 & 0.18 
         &  & 3.80 & 1.15 & 14.65 \\
    4  & & \cellcolor{cayenne!30}0.01 & \cellcolor{cayenne!30}0.03 & \cellcolor{cayenne!30}0.08 
         &  & \cellcolor{cayenne!30}0.68 & \cellcolor{cayenne!30}1.01 & \cellcolor{cayenne!30}0.56 \\
      5 &  & 0.01 & 0.04 & 0.09 
         &  & 0.70 & 1.05 & 0.58 \\
      6 & & 0.02 & 0.05 & 0.10 
         &  & 1.72 & 1.20 & 0.65 \\
    \cline{2-9}
    2 & \multirow{5}{*}{\textbf{Pos Err}}
    & 0.45 & 0.56 & - 
         & \multirow{5}{*}{\textbf{CD$_{\text{movable}}$}}
         & 4.20 & 5.30 & 13.00 \\
    3 & & 0.22 & 0.23 & - 
         &  & 1.12 & 0.48 & 12.40 \\
     4  & & \cellcolor{cayenne!30}0.00 & \cellcolor{cayenne!30}0.01 & - 
         &  & \cellcolor{cayenne!30}0.07 & \cellcolor{cayenne!30}0.11 & \cellcolor{cayenne!30}1.95 \\
     5 & & 0.01 & 0.02 & - 
         &  & 0.28 & 0.22 & 2.45 \\
     6 &  & 0.02 & 0.03 & - 
         &  & 0.39 & 0.34 & 2.70 \\
    \cline{2-9}
    2 & \multirow{5}{*}{\textbf{Motion Err}}
      & 0.40 & 0.65 & 0.46 
         & \multirow{5}{*}{\textbf{CD$_{\text{whole}}$}}
         & 4.10 & 7.30 & 6.90 \\
    3 &  & 0.45 & 0.32 & 0.23 
         &  & 1.95 & 1.62 & 2.60 \\
    4 &  & \cellcolor{cayenne!30}0.02 & \cellcolor{cayenne!30}0.18 & \cellcolor{cayenne!30}0.00 
         &  & \cellcolor{cayenne!30}0.80 & \cellcolor{cayenne!30}0.95 & \cellcolor{cayenne!30}0.51 \\
    5 & & 0.03 & 0.19 & 0.01 
         &  & 1.12 & 1.27 & 0.93 \\
    6 & & 0.04 & 0.20 & 0.02 
         &  & 2.84 & 1.99 & 1.55 \\
    \bottomrule
    \end{tabular}
    }
\end{table}

\begin{table}[t]
    \centering
    \caption{\textbf{Sensitivity of repel point count ($N_R$).}
    Lower ($\downarrow$) is better.}
    \vspace{-0.3cm}
    \label{tab:repel_num_ablation}
    \resizebox{0.95\linewidth}{!}{
    \tiny
    \begin{tabular}{cccc}
    \toprule
    \textbf{Metric} 
        & \textbf{$N_R\!=\!500$} 
        & \textbf{$N_R\!=\!2000$} 
        & \textbf{$N_R\!=\!4000$} \\
    \midrule
    \textbf{Ang Err}     
        & 0.11 & 0.11 & 0.12 \\
    \textbf{Pos Err}      
        & 0.01 & 0.01 & 0.01 \\
    \textbf{Motion Err}    
        & 0.57 & 0.55 & 0.58 \\
    \textbf{CD\textsubscript{whole}}   
        & 0.63 & 0.63 & 0.64 \\
    \bottomrule
    \end{tabular}
    }
    \vspace{-0.3cm}
\end{table}

\noindent\textbf{Differentiability of Repulsion Forces.}
The repulsion update $\boldsymbol{\mu}_i^{k, (t)} \leftarrow \boldsymbol{\mu}_i^{k, (t)} + \mathbf{F}_{\text{repel}, i}^k$ is implemented as a fully differentiable operation within the optimization pipeline. The displacement caused by $\mathbf{F}_{\text{repel}, i}^k$ participates directly in the computation graph rather than acting as a post-processing step. Consequently, during backpropagation, gradients flow through the repulsion force term to the transformation parameters $T_k=(\mathbf{R}_k, \mathbf{t}_k)$. This effectively penalizes configurations where the optimization would otherwise drive Gaussians into repulsion zones, encouraging the learning of collision-free trajectories that naturally avoid repel points while satisfying the alignment loss $\mathcal{L}_{\text{art}}$.

\noindent\textbf{Stability and Force Clamping.}
The inverse cubic falloff defined in the main paper ($1/\|\mathbf{r}_j - \boldsymbol{\mu}^k_i\|^3$) provides strong localized gradients but poses a risk of numerical instability (gradient explosion) as the distance approaches zero. To ensure training stability, we implement two specific safeguards: \textbf{(1) Distance Clamping:} We impose a lower bound on the distance denominator. The L2 distance $\|\mathbf{r}_j - \boldsymbol{\mu}^k_i\|_2$ is clipped to a minimum value $\epsilon\!=\!10^{-5}$. This prevents division by zero and bounds the maximum repulsive force applied to any single Gaussian. \textbf{(2) Force Magnitude Saturation:} We further limit the norm of the total force vector $\|\mathbf{F}_{\text{repel}, i}^k\|$ to a maximum threshold $\tau_{\text{max}}$ to prevent outliers from destabilizing the transformation updates in a single iteration. Thus, the effective robust force calculation is given by:
\begin{equation}
    \mathbf{F}^k_{\text{repel}, i} = \text{clip}\left(\sum\limits_{\mathbf{r}_j \in \mathcal{R}} k_r \cdot \frac{(\mathbf{r}_j - \boldsymbol{\mu}^k_i)}{\max\left(\|\mathbf{r}_j - \boldsymbol{\mu}^k_i\|, \epsilon\right)^3}, \tau_{\text{max}}\right),
    \label{eq:repel_clamped}
\end{equation}
where $\text{clip}(\mathbf{v}, \tau_{\text{max}}) = \mathbf{v} \cdot \min(1, \tau_{\text{max}} / \|\mathbf{v}\|)$ denotes the vector magnitude clipping operation.

\begin{table}[t]
    \centering
    \caption{\textbf{Canonical initialization ablation.}
    Lower ($\downarrow$) is better across all metrics.  
    \colorbox{cayenne!30}{\phantom{\rule{1ex}{1ex}}} highlights best-performing strategy.}
    \vspace{-0.3cm}
    \label{tab:beta_ablation}
    \resizebox{\linewidth}{!}{
    \tiny
    \begin{tabular}{p{1.8cm} p{0.7cm} c c | p{0.7cm} c c}
    \toprule
      \raisebox{1.5ex}{\textbf{Strategy}} & \raisebox{1.5ex}{\textbf{Metrics}} &
    \shortstack{\textbf{Table}\\\textbf{(5 parts)}} &
    \shortstack{\textbf{Storage}\\\textbf{(7 parts)}} &
     \raisebox{1.5ex}{\textbf{Metrics}} &
    \shortstack{\textbf{Table}\\\textbf{(5 parts)}} &
    \shortstack{\textbf{Storage}\\\textbf{(7 parts)}} \\
    \hline

     Uniform Interpolation & \multirow{3}{*}{\textbf{Ang Err}}
        & 0.15 & 0.21 
        & \multirow{3}{*}{\textbf{CD$_{\text{static}}$}} & 1.40 & 1.75 \\
    Motion-Aware Per-Part $\beta$ & & 0.12 & 0.18 
        &  & 1.32 & 1.60 \\
    Motion-Aware Global $\beta$\setlength{\abovedisplayskip}{7pt}&  & \cellcolor{cayenne!30}0.03 & \cellcolor{cayenne!30}0.11 
        &  & \cellcolor{cayenne!30}1.18 & \cellcolor{cayenne!30}0.61 \\
    \hline

     Uniform Interpolation & \multirow{3}{*}{\textbf{Motion Err}}
        & 0.30 & 0.70
        & \multirow{3}{*}{\textbf{CD$_{\text{movable}}$}} & 2.40 & 4.20 \\
    Motion-Aware Per-Part $\beta$ & & 0.20 & 0.52
        &  & 2.15 & 3.00 \\
    Motion-Aware Global $\beta$\setlength{\abovedisplayskip}{7pt}& & \cellcolor{cayenne!30}0.01 & \cellcolor{cayenne!30}0.55
        &  & \cellcolor{cayenne!30}1.85 & \cellcolor{cayenne!30}1.83 \\
    \hline

     Uniform Interpolation & \multirow{3}{*}{\textbf{Pos Err}}
        & 0.08 & 0.12
        & \multirow{3}{*}{\textbf{CD$_{\text{whole}}$}} & 1.20 & 1.45 \\
    Motion-Aware Per-Part $\beta$ & & 0.05 & 0.09
        &  & 1.13 & 1.38 \\
    Motion-Aware Global $\beta$\setlength{\abovedisplayskip}{7pt}&  & \cellcolor{cayenne!30}0.00 & \cellcolor{cayenne!30}0.01
        &  & \cellcolor{cayenne!30}1.10 & \cellcolor{cayenne!30}0.63 \\
    \bottomrule
    \end{tabular}
    }
    \vspace{-0.2cm}
\end{table}

\noindent \textbf{Global vs. Per-part Interpolation Weighting.}
As described in \Cref{sec:part}, the interpolation weight $\beta$ is computed once per object from the global motion richness scores $D_{0\rightarrow1}$ and $D_{1\rightarrow0}$. While this scalar coefficient is shared across all matched Gaussians, we find in practice that a global $\beta$ is sufficient for the purpose of initializing a stable canonical field. 
This is because $\beta$ is used only during initialization to place the canonical Gaussians in a reasonable configuration before the full SE(3)-based deformation module is optimized. Once training begins, each Gaussian’s part membership, transformation, and geometry are updated independently, allowing the model to account for heterogeneous motion magnitudes across parts.\looseness-1

We additionally experiment with (i) uniform averaging and (ii) motion-aware per-part $\beta$. As shown in \Cref{tab:beta_ablation}, both alternatives introduce instability and degrade performance. Per-part $\beta$ is especially sensitive to local displacement noise and fails to reflect the actual articulation structure. In contrast, a single global $\beta$ provides a simple and noise-robust prior while keeping the initialization lightweight.\looseness-1

\noindent\textbf{Hyperparameters.}
For loss weighting, we set $\lambda_{\text{part}}{=}0.1$, $\lambda_{\text{art}}{=}1.0$, and $\lambda_{\text{phys}}{=}0.5$, with equal weights across the three physical regularizers. We set the maximum number of parts K according to category-level priors, typically 3–7. 
The repulsion strength is fixed to $k_r{=}5{\times}10^{-4}$, and we sample $N_R{=}2000$ repel points from regions where canonical Gaussians of movable and static parts fall within a $1.5$\ unit length proximity threshold. 
Repel points remain fixed throughout training. 
The SE(3) transformations for each part are optimized jointly with Gaussian parameters using Adam with learning rate $1\mathrm{e}{-3}$. The canonical Gaussian initialization from the two observed states uses 30k iterations of single-state 3DGS followed by 5k iterations of canonical fusion with the global $\beta$ weighting. 

\begin{figure}[t!]
  \centering
  \includegraphics[width=0.99\columnwidth]{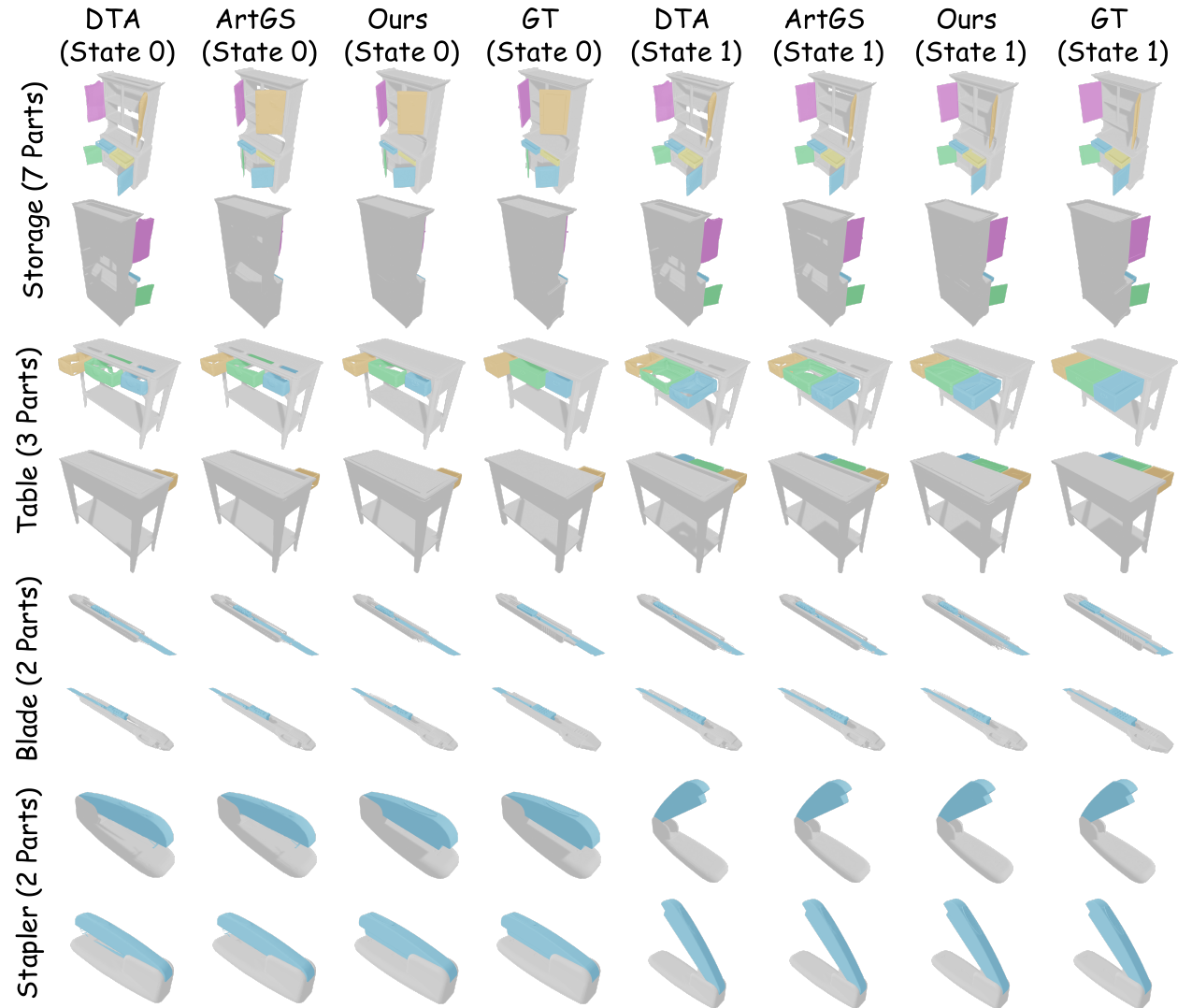}
  \vspace{-0.2cm}
  \caption{\textbf{Mesh visualizations}, confirming high-quality surface reconstruction and consistent part articulation.}
  \label{fig:mesh_results}
  \vspace{-0.3cm}
\end{figure}

\section{Additional Qualitative Examples}
\label{app:qualitative}

\begin{table*}[h]
  \centering
  \setlength\extrarowheight{1pt}
  \caption{\textbf{Inference time for simple and complex objects.} Simple objects have one movable part while complex objects have multiple, denoted by their subscript (\eg Table$_4$ has a static base and three movable parts). 
    \colorbox{cayenne!30}{\phantom{\rule{1ex}{1ex}}} highlights best performing results.}
  \label{tab:inferenceT}
  \vspace{-0.3cm}
  \resizebox{\textwidth}{!}{%
    \begin{tabular}{*{3}{c}*{10}{c}*{7}{c}}
      \toprule
      & \multirow{2}{*}{\textbf{Metric}}
      & \multirow{2}{*}{\textbf{Method}}
        & \multicolumn{10}{c}{\textbf{Simple Objects}}
        & \multicolumn{7}{c}{\textbf{Complex Objects}} \\
      \cmidrule(lr{0.5em}){4-13} \cmidrule(lr{0.5em}){14-20}
      & & &
        \textbf{Foldchair} & \textbf{Fridge} & \textbf{Laptop} & \textbf{Oven}
        & \textbf{Scissor}  & \textbf{Stapler}  & \textbf{USB}     & \textbf{Washer}
        & \textbf{Blade}    & \textbf{Storage} &  
        \textbf{Fridge$_3$}      & \textbf{Table$_4$}     & \textbf{Table$_5$}      & \textbf{Storage$_3$}
        & \textbf{Storage$_4$}     & \textbf{Storage$_7$}     & \textbf{Oven$_4$}     \\ 
      \midrule

      & \multirow{3}{*}{\shortstack{\textbf{Time}\\\textbf{(Min)}}}
      & \textbf{DTA}
        & 29 & 30 & 31 & 29 & 28 & 29 & 31 & 28 & 27 & 28  
        & 32 & 34 & 37 & 32 & 35 & 45 & 35 \\

      & & \textbf{ArtGS}
        & 9 & \cellcolor{cayenne!30}8 & \cellcolor{cayenne!30}7 & \cellcolor{cayenne!30}7
        & \cellcolor{cayenne!30}7 & \cellcolor{cayenne!30}7 & \cellcolor{cayenne!30}7 & \cellcolor{cayenne!30}8 & \cellcolor{cayenne!30}7 & \cellcolor{cayenne!30}8
        & \cellcolor{cayenne!30}8 & \cellcolor{cayenne!30}8 & \cellcolor{cayenne!30}8 & \cellcolor{cayenne!30}8 & \cellcolor{cayenne!30}8 & \cellcolor{cayenne!30}8 & \cellcolor{cayenne!30}8 \\

      & & \shortstack{\modelname{}}
        & \cellcolor{cayenne!30}8 & 9 & \cellcolor{cayenne!30}7 & 8
        & \cellcolor{cayenne!30}7 & 8 & \cellcolor{cayenne!30}7 & \cellcolor{cayenne!30}8
        & \cellcolor{cayenne!30}7 & 9
        & 9 & \cellcolor{cayenne!30}8 & 9 & \cellcolor{cayenne!30}8
        & 9 & 10 & 9 \\
      \bottomrule
    \end{tabular}%
  }
\end{table*}
\begin{table*}[t!]
    \centering
    \scriptsize
    \setlength{\tabcolsep}{2pt}
    \caption{\textbf{\modelnamenc{} module removal ablations} on the two most complex objects in our evaluation, {Table (5 parts)} and {Storage (7 parts)}.  Lower ($\downarrow$) is better on all metrics. \colorbox{cayenne!30}{\phantom{\rule{1ex}{1ex}}} shows results with {all \modelnamenc{} modules} while \colorbox{espresso!15}{\phantom{\rule{1ex}{1ex}}} highlights severe failures by removing components of our method. Severe failures are defined as metrics that are more than 5 times worse than the full \modelnamenc{} for the same object.}
    \label{tab:main_ablation}
    \vspace{-0.3cm}
    \begin{tabularx}{0.95\textwidth}{c c *{6}{>{\centering\arraybackslash}X}}
         \toprule
         \textbf{Objects} & \textbf{Methods} & \textbf{AngErr} & \textbf{PosErr} & \textbf{MotionErr} &  \textbf{CD\textsubscript{static}} &  \textbf{CD\textsubscript{movable}} &  \textbf{CD\textsubscript{whole}} \\
        \cmidrule{1-8}
        \multirow{5}{*}{{\begin{tabular}{c} \textbf{Table} \\ \textbf{(5 parts)} \end{tabular}}} 
            & \xmark~~\textbf{part parameters}      & 0.21 & \cellcolor{espresso!15} 0.08 & \cellcolor{espresso!15} 7.32 & \cellcolor{espresso!15} 7.35 & \cellcolor{espresso!15}  145.17 & 3.10  \\
            & \xmark~~\textbf{repel points}           & 0.09 & 0.16 & \cellcolor{espresso!15} 
 0.48 & 1.19 & 4.82 & 1.85 \\
            & \xmark~~\textbf{physical constraints}   & 0.05 & 0.03 & \cellcolor{espresso!15} 
 0.18 & 1.32 & 4.47 & 1.65  \\
            & \xmark~~\textbf{canonical init}         & 0.14 & \cellcolor{espresso!15} 0.06 & \cellcolor{espresso!15} 
 6.32 & 2.47 & \cellcolor{espresso!15}  117.25 & 2.62 \\
            & \modelname{} \textbf{(all)}                      & \cellcolor{cayenne!30} 0.03 & \cellcolor{cayenne!30} 0.00 & \cellcolor{cayenne!30} 0.01 & \cellcolor{cayenne!30} 1.18 & \cellcolor{cayenne!30} 1.85 & \cellcolor{cayenne!30} 1.10 \\
        \hline
        \multirow{5}{*}{{\begin{tabular}{c} \textbf{Storage} \\ \textbf{(7 parts)} \end{tabular}}} 
            & \xmark~~\textbf{part parameters}      & 0.26 &\cellcolor{espresso!15} 0.11 & \cellcolor{espresso!15} 
            10.43 & 2.95 & \cellcolor{espresso!15} 198.67 & 3.54 \\
            & \xmark~~\textbf{repel points}           & 0.16 & 0.14 & 1.32 & 0.93 & \cellcolor{espresso!15} 7.43 & 2.04 \\
            & \xmark~~\textbf{physical constraints}   & 0.04 & \cellcolor{espresso!15} 0.05 & 0.04 & 1.22 & 4.54 & 1.12 \\
            & \xmark~~\textbf{canonical init}         & \cellcolor{espresso!15}  22.15 & \cellcolor{espresso!15} 0.93 & \cellcolor{espresso!15}  19.67 & 0.79 & \cellcolor{espresso!15}  442.32 & 1.89 \\
            & \modelname{} \textbf{(all)}                      & \cellcolor{cayenne!30} 0.11 & \cellcolor{cayenne!30} 0.01 & \cellcolor{cayenne!30} 0.55 & \cellcolor{cayenne!30} 0.61 & \cellcolor{cayenne!30} 1.83 & \cellcolor{cayenne!30} 0.63 \\
        \hline
    \end{tabularx}
    \vspace{-0.1cm}
\end{table*}

\begin{figure}[t!]
  \centering
  \includegraphics[width=0.99\columnwidth]{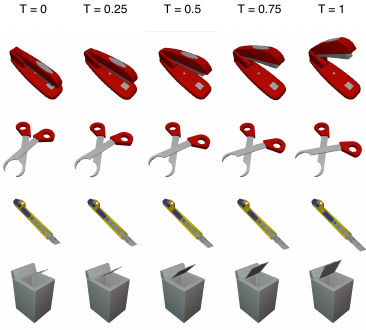}
  \vspace{-0.4cm}
  \caption{\textbf{\modelnamenc{} qualitative results} on 2-part objects with different joints and distinct geometry structures.}
  \label{fig:qualitative_results_simple}
  \vspace{-0.3cm}
\end{figure}

\textbf{Mesh Visualization.} \Cref{fig:mesh_results} shows qualitative comparisons across four articulated objects, \ie Storage (7 Parts), Table (3 Parts), Blade (2 Parts), and Stapler (2 Parts), under State 0 and State 1. Overall, \modelnamenc{} closely matches GT in both geometry and articulation consistency across states. The improvements are especially visible for the multi-part Storage (7 Parts) and Table (3 Parts) examples.

\noindent\textbf{Motion Trajectory Visualization.} \Cref{fig:qualitative_results_simple} presents additional 2-part objects exhibiting diverse geometries and joint types, including rotary (scissors), prismatic (utility knife), and hinged motion (stapler, container lid). Across all examples, \modelnamenc{} produces smooth and monotonically consistent motion trajectories as the articulation parameter T progresses from 0 to 1. The movable parts follow realistic kinematic paths without drifting, collapsing into the static base, or introducing geometric distortion. Notably, fine-scale geometry such as the scissor blades and the tapered cutter head remains stable throughout the motion sequence, demonstrating the robustness of our method.

\section{Inference Time} 
Table \ref{tab:inferenceT} compares the per-object inference runtimes of DTA, ArtGS, and our method \modelnamenc{} on both simple (one movable part) and complex (multiple movable parts) objects. On the ten simple objects, DTA requires between 28 and 31 minutes each, whereas both ArtGS and \modelnamenc{} complete inference in under 10 minutes, yielding roughly a 70–75\% speedup. Notably, \modelnamenc{} achieves the best or tied-best time on eight out of ten simple objects, with ArtGS holding a 1min edge only on Fridge and Stapler.  Despite incorporating additional part-awareness and physical constraints, our method still matches ArtGS’s 8-minute inference performance on most complex objects (and only modestly increases to 10 minutes on the highest-complexity case, Storage$_7$). Overall, \modelnamenc{} delivers state-of-the-art efficiency even with its extra inferential overhead.

\section{Additional Ablations}

\subsection{Sensitivity Ablation}
In \Cref{tab:main_ablation}, we further perform a module-removal ablation to quantify the sensitivity of \modelnamenc{} to each design component. Starting from the full \modelnamenc{} model, we sequentially disable part parameters, repel points, physical constraints, and canonical initialization.

Removing the \textbf{part parameters} leads to the most severe (three orders of magnitude) degradation across both objects. MotionErr increases by more than $700\times$ (0.01$\rightarrow$7.32) and $\text{CD}_{\text{movable}}$ by $\sim$78$\times$ (1.85$\rightarrow$145.17) on the 5-part {Table} object. 
On the 7-part {Storage} object, MotionErr rises $\sim$19$\times$ (0.55$\rightarrow$10.43) and $\text{CD}_{\text{movable}}$ increases over $100\times$ (1.83$\rightarrow$198.67). 
Angular and motion errors spike dramatically (\eg Ang Err from 0.03 to 0.21 and Motion Err from 0.01 to 7.32 on the Table object), while $\text{CD}_{\text{movable}}$ skyrockets by over 70$\times$. This confirms that semantic part disentanglement is essential for stable articulation and coherent geometry recovery. Without explicit part identity supervision, the model fails to isolate and track distinct motions, leading to collapsed or entangled reconstructions.

\begin{table*}[t!]
    \centering
    \scriptsize
    \setlength{\tabcolsep}{2pt}
    \caption{\textbf{Ablations on physics-informed regularization}, on the two most complex objects in our evaluation, {Table (5 parts)} and {Storage (7 parts)}. Lower ($\downarrow$) is better on all metrics. \colorbox{cayenne!30}{\phantom{\rule{1ex}{1ex}}} highlights the best results.}
    \label{tab:pc_ablation}
    \vspace{-0.3cm}
    \begin{tabularx}{0.99\textwidth}{c c *{6}{>{\centering\arraybackslash}X}}
        \toprule
        \textbf{Objects} & \textbf{Methods} &
        \textbf{AngErr} & \textbf{PosErr} & \textbf{MotionErr} &
        \textbf{CD{\textsubscript{static}}} &  \textbf{CD{\textsubscript{movable}}} & \textbf{CD{\textsubscript{whole}}} \\
        \cmidrule{1-8}
        \multirow{4}{*}{{\begin{tabular}{c}\textbf{Table}\\\textbf{(5 parts)}\end{tabular}}}
            & \textbf{no physical constraints} & 0.05 & 0.03 & 0.18 & 1.32 & 4.47 & 1.65 \\
            & +~\textbf{contact loss} & 0.05 & 0.02 & 0.17 & \cellcolor{cayenne!30} 1.18 & \cellcolor{cayenne!30} 1.78 & \cellcolor{cayenne!30} 1.22 \\
            & +~\textbf{velocity consistency} & \cellcolor{cayenne!30}0.03 & 0.01 & 0.02 & 1.33 & 3.11 & 1.52 \\
            & +~\textbf{vector-field alignment} (\modelname{}) &
            \cellcolor{cayenne!30}0.03 & \cellcolor{cayenne!30}0.00 & \cellcolor{cayenne!30}0.01 &
            1.22 & 2.22 & 1.41 \\
        \hline
        \multirow{4}{*}{{\begin{tabular}{c}\textbf{Storage}\\\textbf{(7 parts)}\end{tabular}}}
            & \textbf{no physical constraints} & 0.04 & 0.05 & \cellcolor{cayenne!30}0.04 & 1.22 & 4.54 & 1.12 \\
            & +~\textbf{contact loss} & 0.05 & \cellcolor{cayenne!30}0.04 & \cellcolor{cayenne!30}0.04 & \cellcolor{cayenne!30}0.96 & \cellcolor{cayenne!30}2.12 & 0.74 \\
            & +~\textbf{velocity consistency} & 0.06 & \cellcolor{cayenne!30}0.04 & \cellcolor{cayenne!30}0.04 & 1.21 & 4.01 & \cellcolor{cayenne!30}0.62 \\
            & +~\textbf{vector-field alignment} (\modelname{}) &
            \cellcolor{cayenne!30}0.03 & \cellcolor{cayenne!30}0.04 & \cellcolor{cayenne!30}0.04 &
            1.22 & 3.56 & 0.71 \\
        \bottomrule
    \end{tabularx}
    \vspace{-0.2cm}
\end{table*}
\begin{table}[t!]
    \centering
     \setlength\extrarowheight{1pt}
    \scriptsize
    \setlength{\tabcolsep}{2pt}
    \caption{\textbf{\modelnamenc{} performance by transformation type.} 
    Evaluation across objects undergoing {only translation} or {only rotation} motions. 
    Lower ($\downarrow$) is better for all metrics.}
    \label{tab:trans_rotate}
    \vspace{-0.3cm}
      \resizebox{0.99\linewidth}{!}{%
    \begin{tabularx}{\columnwidth}{c c             
    *{3}{>{\centering\arraybackslash}m{0.075\columnwidth}}  
    *{3}{>{\centering\arraybackslash}m{0.115\columnwidth}}
}
        \toprule
        \textbf{Category} & \textbf{Objects} &
        \textbf{Ang} & \textbf{Pos} & \textbf{Motion} &
        \textbf{CD{\textsubscript{static}}} & $ \textbf{CD{\textsubscript{movable}}}$ & $ \textbf{CD{\textsubscript{whole}}}$ \\
        \cmidrule{1-8}
        \multirow{4}{*}{\rotatebox{90}{\textbf{Translation}}} 
            & Blade (2 parts) & 0.01 & -- & 0.00 & 0.03 & 0.06 & 0.04 \\
            & Storage (2 parts) & 0.01 & -- & 0.00 & 0.04 & 0.04 & 0.04 \\
            & Table (5 parts) & 0.03 & -- & 0.00 & 0.56 & 1.95 & 0.51 \\
            & \textbf{Average} & 0.02 & -- & 0.00 & 0.21 & 0.68 & 0.20 \\
        \hline
        \multirow{4}{*}{\rotatebox{90}{\textbf{Rotation}}}
            & Laptop (2 parts) & 0.01 & 0.00 & 0.01 & 0.07 & 0.09 & 0.08 \\
            & Fridge (3 parts) & 0.01 & 0.00 & 0.02 & 0.59 & 0.08 & 0.73 \\
            & Oven (4 parts) & 0.03 & 0.01 & 0.18 & 1.01 & 0.11 & 0.95 \\
            & \textbf{Average} & 0.02 & 0.00 & 0.07 & 0.56 & 0.09 & 0.59 \\
        \bottomrule
    \end{tabularx}
    }
    \vspace{-0.2cm}
\end{table}

Disabling the \textbf{repel points} has a noticeable effect on motion accuracy but limited influence on geometry quality. On the Table object, motion error increases nearly 50$\times$ (from 0.01 to 0.48), while angular and positional errors also rise, suggesting that the lack of inter-part repulsion leads to ambiguity in part-specific transformations. However, $\text{CD}_{\text{whole}}$ remains relatively stable, confirming that the Gaussian reconstruction itself is unaffected. 

The \textbf{physical constraints} contribute moderate improvements, particularly in reducing $\text{CD}_{\text{movable}}$ and motion error. On both objects, removing these constraints leads to visible but not catastrophic performance drops (\eg Pos Err from 0.01 to 0.05 and $\text{CD}_{\text{movable}}$ from 1.83 to 4.54 on Storage), indicating that they provide useful geometric regularization but are not the sole factor in driving accuracy.

Finally, removing \textbf{canonical initialization} results in the most unstable training behavior. Angular error explodes from 0.11 to 22.15 on Storage, and motion error increases by over 35$\times$ on both objects. Results highlight the importance of starting from a stable, geometry-aligned canonical state to enable robust part tracking and learning. 

\subsection{Ablation on Physics-Informed Losses}
We additionally perform ablations to quantify the impact of each physical constraint. As shown in \Cref{tab:pc_ablation}, each physical loss meaningfully contributes to improved motion accuracy and geometry quality. \textbf{Contact loss} leads to the largest drop in geometry errors. For instance, on the Table object, which exhibits multi-axis, rotational articulation, contact loss cuts $\text{CD}_\text{movable}$ by more than half (4.47$\rightarrow$1.78) and $\text{CD}_{\text{whole}}$ by {26\%} (1.65$\rightarrow$1.22), indicating far less interpenetration and more realistic results. 
\textbf{Velocity consistency} improves motion quality, nearly eliminating motion errors (\eg reducing Motion Err from 0.18 to 0.02). \textbf{Vector-field alignment} yields the lowest angular and positional errors, driving errors down across the board and yielding the most physically plausible, accurate articulations overall. These results demonstrate that the proposed physical constraints act in complementary ways to enable physically plausible, precise articulation and geometry reconstruction.
{Storage (7 parts)} shows reduced inter-part penetration ($\text{CD}_{\text{movable}}$:~4.54$\rightarrow$2.12, $\text{CD}_{\text{whole}}$:~1.12$\rightarrow$0.74), while motion errors remain nearly unchanged ($\text{MotionErr}=0.04$). 
Here, the baseline motion is already simple and prismatic, so the constraints primarily enforce geometric separation rather than further reducing dynamic error. 
Overall, these results indicate that the proposed constraints provide a consistent and interpretable improvement in both physical plausibility and geometric fidelity, particularly for complex, multi-axis articulations.

\subsection{Translation vs. Rotation Ablation} We provide an ablation analysis for translation-only and rotation-only objects. \Cref{tab:trans_rotate} results show that \modelnamenc{} achieves consistently low error across both motion types. Notably, objects with pure translation exhibit near-zero motion errors and lower average CD metrics, reflecting the relative simplicity of prismatic articulation. Rotational objects maintain low error as well, but with slightly higher averages due to increased articulation complexity. We also observe that rotational objects tend to have higher CD values compared to translational objects (\eg Avg. $\text{CD}_\text{whole}$: 0.59 vs. 0.20), likely due to increased geometric complexity.

\begin{table}[t]
    \centering
    \caption{\textbf{Robustness to noisy repel-point initialization.} Lower ($\downarrow$) is better on all metrics. \colorbox{cayenne!30}{\phantom{\rule{1ex}{1ex}}} highlights the best results.}
    \vspace{-0.4cm}
    \label{tab:repel_noise_robustness}
    \resizebox{0.99\columnwidth}{!}{%
        \begin{tabular}{c c  c c c c c c}
        \toprule
        \multirow{2}{*}{\textbf{Metric}} 
        & \multirow{2}{*}{$\boldsymbol{\sigma_r}$} 
        & {\begin{tabular}{c} \textbf{Foldchair} \\ \textbf{(2 parts)} \end{tabular}}
        & {\begin{tabular}{c} \textbf{Stapler} \\ \textbf{(2 parts)} \end{tabular}}
        & {\begin{tabular}{c} \textbf{Blade} \\ \textbf{(2 parts)} \end{tabular}}
        & {\begin{tabular}{c} \textbf{Oven} \\ \textbf{(4 parts)} \end{tabular}}
        & {\begin{tabular}{c} \textbf{Table} \\ \textbf{(5 parts)} \end{tabular}}
        & {\begin{tabular}{c} \textbf{Storage} \\ \textbf{(7 parts)} \end{tabular}} \\
        \hline
        
        \multirow{4}{*}{\shortstack{\textbf{Ang}\\\textbf{Err} $\downarrow$}} 
        & 0.00 & \cellcolor{cayenne!30} 0.01 & \cellcolor{cayenne!30} 0.01 & \cellcolor{cayenne!30} 0.01 & \cellcolor{cayenne!30} 0.03 & \cellcolor{cayenne!30} 0.30 & \cellcolor{cayenne!30} 0.11 \\
        & 0.01 & \cellcolor{cayenne!30} 0.01 & 0.02 & \cellcolor{cayenne!30} 0.01 & 0.04 & 0.31 & 0.12 \\
        & 0.03 & 0.02 & 0.03 & 0.02 & 0.06 & 0.34 & 0.14 \\
        & 0.05 & 0.03 & 0.04 & 0.03 & 0.08 & 0.37 & 0.17 \\
        
        \cmidrule(lr){1-8}
        
        \multirow{4}{*}{\shortstack{\textbf{Pos}\\\textbf{Err} $\downarrow$}} 
        & 0.00 & \cellcolor{cayenne!30} 0.00 & \cellcolor{cayenne!30} 0.01 & - & \cellcolor{cayenne!30} 0.01 & \cellcolor{cayenne!30} 0.00 & \cellcolor{cayenne!30} 0.01 \\
        & 0.01 & \cellcolor{cayenne!30} 0.00 & \cellcolor{cayenne!30} 0.01 & - & \cellcolor{cayenne!30} 0.01 & 0.01 & \cellcolor{cayenne!30} 0.01 \\
        & 0.03 & 0.01 & 0.02 & - & 0.02 & 0.02 & 0.02 \\
        & 0.05 & 0.02 & 0.03 & - & 0.03 & 0.03 & 0.03 \\
        
        \cmidrule(lr){1-8}

        \multirow{4}{*}{\shortstack{\textbf{Motion}\\\textbf{Err} $\downarrow$}} 
        & 0.00 & \cellcolor{cayenne!30} 0.01 & \cellcolor{cayenne!30} 0.00 & \cellcolor{cayenne!30} 0.00 & \cellcolor{cayenne!30} 0.18 & \cellcolor{cayenne!30} 0.01 & \cellcolor{cayenne!30} 0.55 \\
        & 0.01 & \cellcolor{cayenne!30} 0.01 & 0.01 & 0.01 & 0.19 & 0.02 & 0.58 \\
        & 0.03 & 0.02 & 0.02 & 0.02 & 0.23 & 0.03 & 0.64 \\
        & 0.05 & 0.03 & 0.03 & 0.03 & 0.28 & 0.04 & 0.72 \\
        
        \hline

        \multirow{4}{*}{\textbf{CD\textsubscript{whole}} $\downarrow$} 
        & 0.00 & \cellcolor{cayenne!30} 0.19 & \cellcolor{cayenne!30} 1.45 & \cellcolor{cayenne!30} 0.35 & \cellcolor{cayenne!30} 0.95 & \cellcolor{cayenne!30} 1.10 & \cellcolor{cayenne!30} 0.63 \\
        & 0.01 & 0.20 & 1.46 & 0.36 & 0.96 & 1.12 & 0.65 \\
        & 0.03 & 0.21 & 1.48 & 0.38 & 0.98 & 1.14 & 0.67 \\
        & 0.05 & 0.23 & 1.51 & 0.40 & 1.00 & 1.18 & 0.71 \\
        
        \bottomrule
    \end{tabular}
    }
        \vspace{-0.2cm}
\end{table}

\subsection{Noisy Repel Points Initialization}
To evaluate sensitivity to repel-point initialization, we perturb the initially generated repel points with small random 3D offsets with magnitude $\sigma_r$ (\eg $\sigma_r\!=\!0.01$ corresponds to $\sim$1\% of the object’s spatial extent). \Cref{tab:repel_noise_robustness} shows performance remains stable under moderate noise.

\subsection{Fixed vs. Dynamic Repel Points}
We compare fixed repel points with a dynamic variant that recomputes them during training. As shown in \Cref{tab:fixed_dynamic_repel_per_object}, the results are nearly identical overall, and dynamic updates provide only minor gains under noisy initialization, confirming that fixed repel points are generally sufficient and already offer a stable choice in practice.

\begin{table}[t]
\centering
\caption{\textbf{Repel points robustness.}
We compare \textbf{Fixed} repel points with a \textbf{Dynamic} variant that refreshes them every $K{=}5$k iterations.
\textbf{Clean Init} uses default repel points; \textbf{Noisy Init} perturbs them before optimization (\eg $\sigma_r{=}0.05$). \colorbox{cayenne!30}{\phantom{\rule{1ex}{1ex}}} highlights best results.}
\vspace{-0.3cm}
\label{tab:fixed_dynamic_repel_per_object}
\resizebox{0.99\columnwidth}{!}{
\begin{tabular}{c c  c c c c c c}
\toprule
\textbf{Metric} & \textbf{Setting} 
& {\begin{tabular}{c}\textbf{Foldchair}\\\textbf{(2 parts)}\end{tabular}}
& {\begin{tabular}{c}\textbf{Stapler}\\\textbf{(2 parts)}\end{tabular}}
& {\begin{tabular}{c}\textbf{Blade}\\\textbf{(2 parts)}\end{tabular}}
& {\begin{tabular}{c}\textbf{Oven}\\\textbf{(4 parts)}\end{tabular}}
& {\begin{tabular}{c}\textbf{Table}\\\textbf{(5 parts)}\end{tabular}}
& {\begin{tabular}{c}\textbf{Storage}\\\textbf{(7 parts)}\end{tabular}} \\
\midrule

\multirow{4}{*}{\shortstack{\textbf{Ang}\\\textbf{Err} $\downarrow$}}
& Clean + Fixed   
& \cellcolor{cayenne!30}0.01 & \cellcolor{cayenne!30}0.01 & \cellcolor{cayenne!30}0.01 & \cellcolor{cayenne!30}0.03 & \cellcolor{cayenne!30}0.30 & \cellcolor{cayenne!30}0.11 \\
& Clean + Dynamic 
& \cellcolor{cayenne!30}0.01 & \cellcolor{cayenne!30}0.01 & \cellcolor{cayenne!30}0.01 & \cellcolor{cayenne!30}0.03 & \cellcolor{cayenne!30}0.30 & \cellcolor{cayenne!30}0.11 \\
& Noisy + Fixed   
& 0.03 & 0.04 & 0.03 & 0.08 & 0.37 & 0.17 \\
& Noisy + Dynamic 
& 0.03 & 0.04 & 0.03 & 0.08 & 0.35 & 0.17 \\

\cmidrule(lr){1-8}

\multirow{4}{*}{\shortstack{\textbf{Pos}\\\textbf{Err} $\downarrow$}}
& Clean + Fixed   
& \cellcolor{cayenne!30}0.00 & \cellcolor{cayenne!30}0.01 & - & \cellcolor{cayenne!30}0.01 & \cellcolor{cayenne!30}0.00 & \cellcolor{cayenne!30}0.01 \\
& Clean + Dynamic 
& \cellcolor{cayenne!30}0.00 & \cellcolor{cayenne!30}0.01 & - & \cellcolor{cayenne!30}0.01 & \cellcolor{cayenne!30}0.00 & \cellcolor{cayenne!30}0.01 \\
& Noisy + Fixed   
& 0.02 & 0.03 & - & 0.03 & 0.03 & 0.03 \\
& Noisy + Dynamic 
& 0.02 & 0.03 & - & 0.03 & 0.02 & 0.03 \\

\cmidrule(lr){1-8}

\multirow{4}{*}{\shortstack{\textbf{Motion}\\\textbf{Err} $\downarrow$}}
& Clean + Fixed   
& \cellcolor{cayenne!30}0.01 & \cellcolor{cayenne!30}0.00 & \cellcolor{cayenne!30}0.00 & \cellcolor{cayenne!30}0.18 & \cellcolor{cayenne!30}0.01 & 0.55 \\
& Clean + Dynamic 
& \cellcolor{cayenne!30}0.01 & \cellcolor{cayenne!30}0.00 & \cellcolor{cayenne!30}0.00 & \cellcolor{cayenne!30}0.18 & \cellcolor{cayenne!30}0.01 & \cellcolor{cayenne!30}0.54 \\
& Noisy + Fixed   
& 0.03 & 0.03 & 0.03 & 0.28 & 0.04 & 0.72 \\
& Noisy + Dynamic 
& 0.03 & 0.03 & 0.03 & 0.26 & 0.04 & 0.69 \\

\midrule

\multirow{4}{*}{\textbf{CD\textsubscript{whole}} $\downarrow$}
& Clean + Fixed   
& \cellcolor{cayenne!30}0.19 & 1.45 & \cellcolor{cayenne!30}0.35 & \cellcolor{cayenne!30}0.95 & 1.10 & \cellcolor{cayenne!30}0.63 \\
& Clean + Dynamic 
& \cellcolor{cayenne!30}0.19 & 1.45 & \cellcolor{cayenne!30}0.35 & \cellcolor{cayenne!30}0.95 & \cellcolor{cayenne!30}1.09 & \cellcolor{cayenne!30}0.63 \\
& Noisy + Fixed   
& 0.23 & 1.51 & 0.40 & 1.00 & 1.18 & 0.71 \\
& Noisy + Dynamic 
& 0.22 & \cellcolor{cayenne!30}1.43 & 0.39 & 0.99 & 1.16 & 0.69 \\

\bottomrule
\end{tabular}
}
\end{table}

\subsection{Part Number (K) Selection} 
We follow standard practice in articulated modeling and set \(K\) to the number of movable parts for fair comparison with prior work, while treating it as an upper bound in practice. Beyond the mis-specification study in \Cref{tab:k_ablation}, we further examine a practically relevant regime of mild over-estimation in \Cref{tab:k_sensitivity}, comparing \(K_{\mathrm{GT}}\) against \(K_{\mathrm{GT}}+2\) and \(K_{\mathrm{GT}}+4\). Results show that \modelnamenc{} remains robust when \(K\) is moderately over-specified, with only small changes in articulation and reconstruction quality.
Using \(K_{\mathrm{GT}}+2\) generally preserves performance across angular error, positional error, motion error, and \(\text{CD}_{\text{whole}}\). For example, on \textsc{Table} and \textsc{Storage}, the whole-object Chamfer Distance changes only from \(1.10 \rightarrow 1.12\) and \(0.63 \rightarrow 0.65\), respectively. Even with \(K_{\mathrm{GT}}+4\), performance degrades only modestly on more complex objects, suggesting that redundant part slots are largely suppressed during optimization rather than causing catastrophic failure.\looseness-1

\begin{table}[t]
\centering
\caption{\textbf{Sensitivity to the number of parts.}
$K_{\text{GT}}$ denotes ground-truth number of parts $K$. \colorbox{cayenne!30}{\phantom{\rule{1ex}{1ex}}} highlights the best results.}
\vspace{-0.3cm}
\label{tab:k_sensitivity}
\resizebox{0.99\columnwidth}{!}{
\begin{tabular}{c c c c c c c c}
\toprule
\textbf{Metric} & \textbf{$K$ Setting}
& {\begin{tabular}{c}\textbf{Foldchair}\\(2 parts)\end{tabular}}
& {\begin{tabular}{c}\textbf{Stapler}\\(2 parts)\end{tabular}}
& {\begin{tabular}{c}\textbf{Blade}\\(2 parts)\end{tabular}}
& {\begin{tabular}{c}\textbf{Oven}\\(4 parts)\end{tabular}}
& {\begin{tabular}{c}\textbf{Table}\\(5 parts)\end{tabular}}
& {\begin{tabular}{c}\textbf{Storage}\\(7 parts)\end{tabular}} \\
\midrule

\multirow{3}{*}{\textbf{Ang Err} $\downarrow$}
& $K_{\text{GT}}$     & \cellcolor{cayenne!30}0.01 & \cellcolor{cayenne!30}0.01 & \cellcolor{cayenne!30}0.01 & \cellcolor{cayenne!30}0.03 & \cellcolor{cayenne!30}0.03 & \cellcolor{cayenne!30}0.11 \\
& $K_{\text{GT}}+2$   & \cellcolor{cayenne!30}0.01 & \cellcolor{cayenne!30}0.01 & \cellcolor{cayenne!30}0.01 & \cellcolor{cayenne!30}0.03 & 0.04 & \cellcolor{cayenne!30}0.11 \\
& $K_{\text{GT}}+4$   & 0.02 & 0.02 & \cellcolor{cayenne!30}0.01 & 0.04 & 0.05 & 0.12 \\
\midrule

\multirow{3}{*}{\textbf{Pos Err} $\downarrow$}
& $K_{\text{GT}}$     & \cellcolor{cayenne!30}0.00 & \cellcolor{cayenne!30}0.01 & - & \cellcolor{cayenne!30}0.01 & \cellcolor{cayenne!30}0.00 & \cellcolor{cayenne!30}0.01 \\
& $K_{\text{GT}}+2$   & \cellcolor{cayenne!30}0.00 & \cellcolor{cayenne!30}0.01 & - & \cellcolor{cayenne!30}0.01 & 0.01 & \cellcolor{cayenne!30}0.01 \\
& $K_{\text{GT}}+4$   & 0.01 & \cellcolor{cayenne!30}0.01 & - & 0.02 & 0.01 & 0.02 \\
\midrule

\multirow{3}{*}{\textbf{Motion Err} $\downarrow$}
& $K_{\text{GT}}$     & \cellcolor{cayenne!30}0.01 & \cellcolor{cayenne!30}0.00 & \cellcolor{cayenne!30}0.00 & \cellcolor{cayenne!30}0.18 & \cellcolor{cayenne!30}0.01 & \cellcolor{cayenne!30}0.55 \\
& $K_{\text{GT}}+2$   & \cellcolor{cayenne!30}0.01 & 0.01 & \cellcolor{cayenne!30}0.00 & 0.19 & 0.02 & 0.57 \\
& $K_{\text{GT}}+4$   & 0.02 & 0.01 & 0.01 & 0.22 & 0.03 & 0.60 \\
\midrule

\multirow{3}{*}{ \textbf{CD{\textsubscript{whole}}} $\downarrow$}
& $K_{\text{GT}}$     & \cellcolor{cayenne!30}0.19 & \cellcolor{cayenne!30}1.45 & \cellcolor{cayenne!30}0.35 & \cellcolor{cayenne!30}0.95 & \cellcolor{cayenne!30}1.10 & \cellcolor{cayenne!30}0.63 \\
& $K_{\text{GT}}+2$   & 0.20 & 1.46 & 0.36 & 0.96 & 1.12 & 0.65 \\
& $K_{\text{GT}}+4$   & 0.22 & 1.49 & 0.38 & 0.99 & 1.15 & 0.68 \\
\bottomrule
\end{tabular}
}
\vspace{-0.2cm}
\end{table}

\subsection{Repel-Force Exponent Ablation}
We employ $\|\mathbf{r}-\boldsymbol{\mu}\|^{3}$ in \Cref{eq:repel} so that the resulting repulsion vector has an inverse-square magnitude, \ie, $\|\mathbf{F}\|\propto 1/d^{2}$ with $d\!=\!\|\mathbf{r}-\boldsymbol{\mu}\|$, while preserving its direction toward the repel point. In \Cref{tab:repel_exponent_ablation}, we ablate the falloff exponent $p$ in $\mathbf{F}\propto(\mathbf{r}-\boldsymbol{\mu})/\|\mathbf{r}-\boldsymbol{\mu}\|^{p}$ and observe that $p\!=\!3$ provides the best trade-off between preventing interpenetration and maintaining accurate motion and geometry.
\begin{table}[t]
  \centering
  \setlength{\tabcolsep}{8.5pt}
  \renewcommand{\arraystretch}{0.9}
  \caption{\textbf{Repel-Force Ablation.}
  Results averaged over all objects.}
  \vspace{-0.3cm}
  \label{tab:repel_exponent_ablation}
  \resizebox{\columnwidth}{!}{
  \begin{tabular}{c c c c}
  \toprule
  \textbf{Exponent $p$} 
  & \textbf{Motion Err} $\downarrow$ 
  & \textbf{CD{\textsubscript{whole}}} $\downarrow$ 
  & Penetration $\downarrow$ \\
  \midrule
  2 & 0.028 & 0.69 & 0.021 \\
  \cellcolor{cayenne!30}3 & \cellcolor{cayenne!30}0.020 & \cellcolor{cayenne!30}0.66 & \cellcolor{cayenne!30}0.009 \\
  4 & 0.023 & 0.67 & 0.012 \\
  \bottomrule
  \end{tabular}
  }
\end{table}

\section{Photometric Evaluation}
We additionally report photometric metrics averaged over both observation states. As shown in \Cref{tab:photometric_eval}, \modelnamenc{} consistently outperforms ArtGS across all objects and all three metrics, indicating more accurate pixel-level reconstruction and improved perceptual quality. These gains are consistent across simpler and more challenging multi-part objects.

\begin{table}[t!]
\centering
\scriptsize
\setlength\extrarowheight{1pt}
\caption{\textbf{Photometric evaluation.}
Metrics averaged over observation states.  \colorbox{cayenne!30}{\phantom{\rule{1ex}{1ex}}} highlights best performing results.}
\vspace{-0.3cm}
\label{tab:photometric_eval}
\resizebox{0.99\columnwidth}{!}{
\begin{tabular}{c c c c c c c c}
\toprule
\textbf{Metric} & \textbf{Method}
& {\begin{tabular}{c}\textbf{Foldchair}\\\textbf{(2 parts)}\end{tabular}}
& {\begin{tabular}{c}\textbf{Stapler}\\\textbf{(2 parts)}\end{tabular}}
& {\begin{tabular}{c}\textbf{Blade}\\\textbf{(2 parts)}\end{tabular}}
& {\begin{tabular}{c}\textbf{Oven}\\\textbf{(4 parts)}\end{tabular}}
& {\begin{tabular}{c}\textbf{Table}\\\textbf{(5 parts)}\end{tabular}}
& {\begin{tabular}{c}\textbf{Storage}\\\textbf{(7 parts)}\end{tabular}} \\
\midrule

\textbf{PSNR} $\uparrow$
& ArtGS  & 32.4 & 33.1 & 31.7 & 30.2 & 29.6 & 28.7 \\
& \modelname   & \cellcolor{cayenne!30}\textbf{33.6} & \cellcolor{cayenne!30}\textbf{34.2} & \cellcolor{cayenne!30}\textbf{32.9} & \cellcolor{cayenne!30}\textbf{31.4} & \cellcolor{cayenne!30}\textbf{30.8} & \cellcolor{cayenne!30}\textbf{29.9} \\
\midrule

\textbf{SSIM} $\uparrow$
& ArtGS  & 0.968 & 0.972 & 0.961 & 0.950 & 0.942 & 0.934 \\
& \modelname   & \cellcolor{cayenne!30}\textbf{0.975} & \cellcolor{cayenne!30}\textbf{0.979} & \cellcolor{cayenne!30}\textbf{0.970} & \cellcolor{cayenne!30}\textbf{0.959} & \cellcolor{cayenne!30}\textbf{0.951} & \cellcolor{cayenne!30}\textbf{0.944} \\
\midrule

\textbf{LPIPS} $\downarrow$
& ArtGS  & 0.041 & 0.039 & 0.047 & 0.058 & 0.066 & 0.072 \\
& \modelname   & \cellcolor{cayenne!30}\textbf{0.035} & \cellcolor{cayenne!30}\textbf{0.033}  & \cellcolor{cayenne!30}\textbf{0.040} & \cellcolor{cayenne!30}\textbf{0.051} & \cellcolor{cayenne!30}\textbf{0.059} & \cellcolor{cayenne!30}\textbf{0.064} \\
\bottomrule
\end{tabular}
}
\vspace{-0.2cm}
\end{table}

\section{Broader Impacts}
The ability to accurately reconstruct and articulate 3D objects has far-reaching implications across robotics, simulation, and digital twin technologies. \modelnamenc{} contributes to this space by enabling precise, physically grounded modeling of complex articulated objects from visual observations. This can facilitate improved interaction and manipulation in embodied agents, enhance simulation fidelity in virtual environments, and support scalable generation of articulated assets for digital content creation, industrial, and educational applications.
While the ability to digitize and manipulate real-world objects raises potential concerns around privacy, intellectual property, or misuse in synthetic media, our model is designed for research and educational use. We encourage responsible deployment practices aligned with consent and attribution norms. Compared to large-scale generative systems, our model is computationally lightweight and environmentally efficient, and we view its benefits in controllable, interpretable object modeling as outweighing its risks when applied ethically.\looseness-1

\end{document}